\documentclass[10pt,journal,compsoc]{IEEEtran}
\usepackage{amsmath,amsfonts}
\usepackage{array}
\usepackage{textcomp}
\usepackage{stfloats}
\usepackage{url}
\usepackage{verbatim}
\usepackage{graphicx}
\usepackage{cite}
\usepackage{amsthm}
\usepackage{subcaption}
\usepackage{hyperref}
\usepackage[numbers]{natbib}  
\hypersetup{colorlinks=true, linkcolor=blue, citecolor=blue, urlcolor=blue}  

\usepackage{booktabs}
\usepackage{multirow}
\usepackage{makecell}
\usepackage{bm}
\usepackage{mathrsfs}
\usepackage{amssymb} 
\usepackage{tikz}
\usepackage[inkscapelatex=false]{svg}

\usepackage[table]{xcolor}
\usepackage{color}

\usepackage[T1]{fontenc}
\usepackage[utf8]{inputenc}

\newcommand{\std}[1]{\text{\scriptsize$\pm$#1}}

\usepackage[ruled,vlined,linesnumbered]{algorithm2e}

\usepackage[switch]{lineno}  

\newif\iflinenumbers
\linenumbersfalse 

\iflinenumbers
\linenumbers
\modulolinenumbers[1] 
\fi

\newtheorem{theorem}{Theorem}[section]  
\newtheorem{lemma}{Lemma}[section]

\newtheorem{definition}{Definition}[section]  
\newtheorem{corollary}{Corollary}[section]  

\begin{document}
	
	\title{Robust Losses from Univariate Base Functions for Noisy-Label Learning}
	
	\author{
		Peng Hu, Jianwei Ma
		\thanks{
			P. Hu is with the School of Mathematics, Harbin Institute of Technology, Harbin, China.
		}
		\thanks{
			J. Ma is with the Institute for Artificial Intelligence and the School of Earth and Space Sciences, Peking University, Beijing, China, and also with the Institute for Artificial Intelligence and the School of Mathematics, Harbin Institute of Technology, Harbin, China.
		}
		\thanks{
			Corresponding author: J. Ma (e-mail: jwm@pku.edu.cn).
		}
	}

	 \IEEEtitleabstractindextext{
	\begin{abstract}
		Learning with noisy labels is a fundamental problem in training reliable deep neural networks. Robust loss functions provide a direct and effective way to mitigate the adverse effects of label noise. However, most existing robust losses are designed directly at the level of the final multiclass objective, which makes it difficult to systematically characterize and extend their robustness properties. In this paper, we propose a general framework that constructs robust multiclass losses from univariate base functions. By defining mapping operators from base functions to multiclass losses, the robustness of the induced losses can be characterized through simple properties of the base functions. We develop two complementary construction schemes, Target Separation and Binary Reduction, corresponding to inter-class independent and inter-class dependent formulations, respectively. For both schemes, we analyze their symmetry and asymmetry properties and derive corresponding sufficient conditions, which provide theoretical criteria for noise-robust loss design. The proposed framework also provides a new route to constructing symmetric losses, serving as a complement to normalization-based symmetric loss designs. Extensive experiments on synthetic and real-world noisy-label benchmarks demonstrate that the proposed losses achieve competitive or superior performance under various noise settings.
	\end{abstract}
	
		}
	
	\maketitle
	
	\section{Introduction}
	\IEEEPARstart{D}{eep} neural networks (DNNs) have achieved remarkable success across a wide range of tasks~\cite{Yann2015Deep}; however, their performance critically depends on the availability of large-scale datasets with accurate annotations~\cite{Bo2021Survey, Devansh2017Closer}. This requirement is often impractical in practice~\cite{Collin2024Weak}, as acquiring accurately labeled data at scale is costly, labor-intensive, or even infeasible. Consequently, label noise is pervasive in many datasets, arising from human annotation errors, automated labeling pipelines, or weak supervision. Noisy labels can severely degrade the generalization performance of deep neural networks, as they tend to eventually memorize corrupted labels during training. To mitigate this issue, a variety of approaches have been proposed, including improved training strategies~\cite{Han2018co, Eran2017Decoupling, Yu2019How, Xu2019LDMI}, label correction~\cite{Lee2018CleanNet, Vahdat2017Toward, Li2017Learning, Veit2017Learninig}, loss correction~\cite{Giorgio2017Making, Han2018Masking, Szegedy2016Rethinking, goldberger2017training}, and robust losses~\cite{Aritra2017Robust, Zhou2023Asymmetric, Zhang2024LTAPL, Wang2025JAL}. Among them, robust losses mitigate overfitting to noisy labels through careful design and are widely adopted for their simplicity and effectiveness.

	In this context, noise tolerance has emerged as a key concept, referring to the ability of learning algorithms to maintain stable performance under label corruption. Theoretical analyses indicate that symmetric loss functions such as mean absolute error (MAE) are inherently noise-tolerant, whereas the widely used cross-entropy (CE) loss lacks this property. However, MAE is known to be difficult to optimize~\cite{Aritra2017Robust}, which has sparked extensive research on designing loss functions that are both noise-robust and converge rapidly. For example, the Active Passive Loss (APL) framework~\cite{Ma2020NCERCE} splits loss functions into active and passive losses and jointly exploits them to improve optimization efficiency and fitting performance. Experimental results demonstrate that incorporating symmetric loss functions into the Active Passive Loss (APL) framework effectively enhances robustness and generalization performance.
	
	Zhou et al.~\cite{Zhou2021Asymmetric, Zhou2023Asymmetric} further proposed asymmetric loss functions (ALFs) to improve noise robustness, offering a new perspective on robust learning. Theoretical analyses show that ALFs require less restrictive conditions than symmetric ones. Following these analyses, they further proposed the Asymmetric Unhinged Loss (AUL), Asymmetric Generalized Cross Entropy (AGCE), and Asymmetric Exponential Loss (AEL). Building on this framework, Asymmetric Mean Squared Error (AMSE)~\cite{Wang2025JAL} and Variation-Bounded Loss (VBL)~\cite{wang2026variation} were subsequently proposed. While existing robust loss functions are primarily developed by directly designing the final objective, our framework takes a different perspective by constructing losses from univariate base functions. This enables a systematic characterization of loss robustness through simple properties of base functions, leading to broad families of noise-robust losses.

	\begin{figure}[htbp]
		\centering
		
		\subfloat{
			\includegraphics[width=0.49\linewidth]{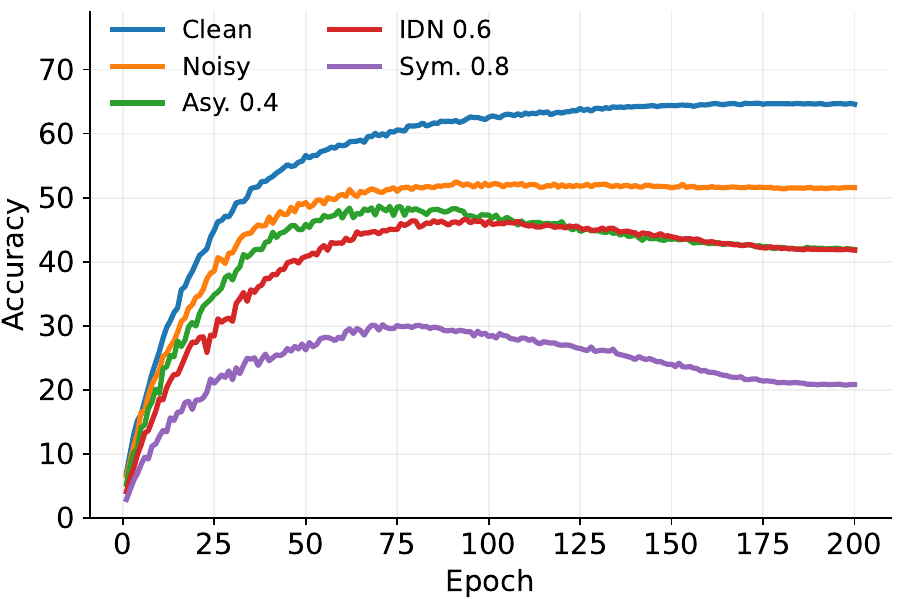}
			\label{fig_GCE}
		}
		\subfloat{
			\includegraphics[width=0.49\linewidth]{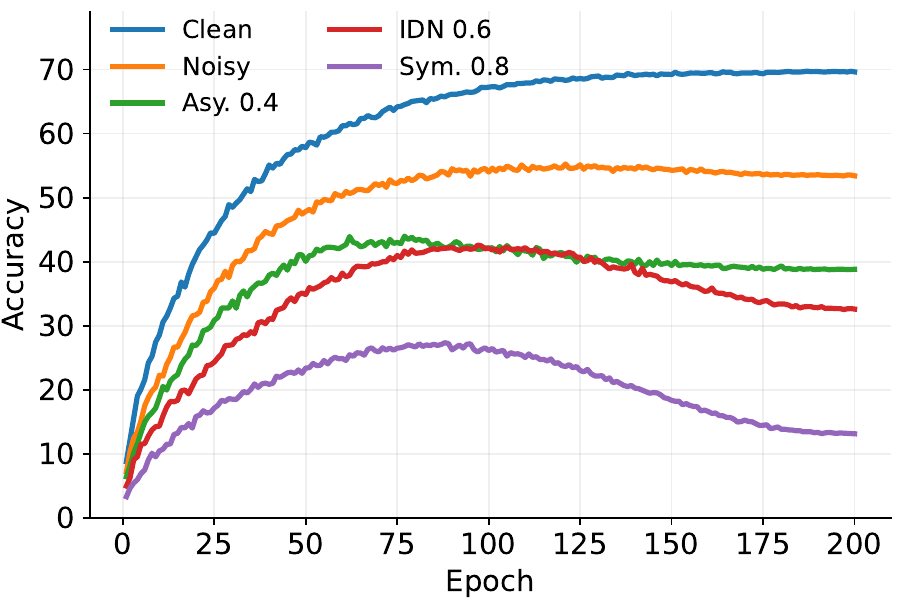}
			\label{fig_NCEAUL}
		}

		\caption{Test accuracy curves of GCE (left) and NCE+AUL (right) on CIFAR-100/CIFAR-100N under five noisy-label settings. Although both methods achieve rapid performance gains during early training, their accuracies deteriorate or saturate in the later stage under several noise conditions, suggesting that the network gradually overfits noisy labels.}
		\label{fig_overfit}
	\end{figure}
	
	\begin{figure*}[htbp]
		\centering
		\includegraphics[width=0.99\linewidth]{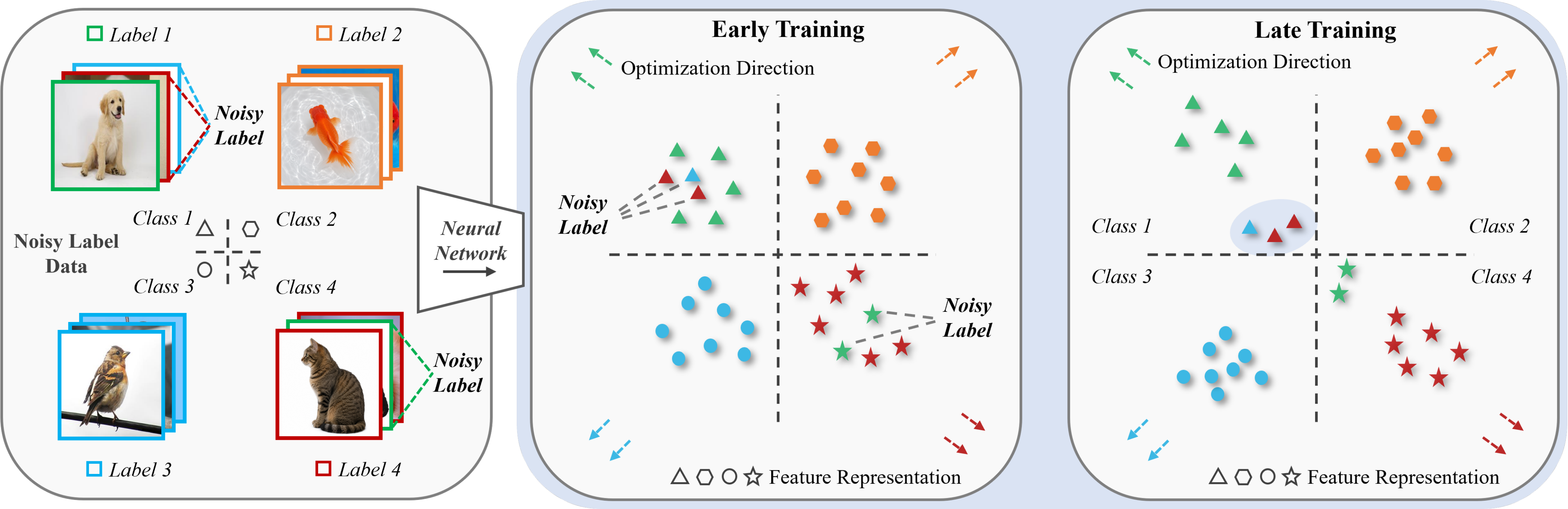}
		\caption{Noisy-Label Learning. Left: original data with noisy labels. Middle: during early training, clean-label samples dominate and help noisy-label samples follow the correct data distribution. Right: during late training, the network overfits noisy labels and separates noisy-label samples from the original data distribution.}
		\label{fig_mov}
	\end{figure*}

	Moreover, it is worth noting that a theoretically well-characterized noise-tolerant function may not necessarily be fully effective in practice. Apart from the optimization challenges of losses such as MAE, many existing robust losses remain vulnerable to overfitting noisy labels. As shown in Fig.~\ref{fig_overfit}, although representative robust losses such as Generalized Cross Entropy (GCE)~\cite{Zhang2018GCE} and Asymmetric Loss Function (NCE+AUL)~\cite{Zhou2021Asymmetric,Zhou2023Asymmetric} achieve rapid performance gains during early training, their accuracies later saturate or even deteriorate under several noisy settings, indicating that the network gradually overfits noisy labels.
	To provide an intuitive explanation of this phenomenon, we illustrate the training dynamics of noisy-label learning in Fig.~\ref{fig_mov}.
	During early training, clean-label samples dominate optimization and guide noisy-label samples toward the correct data distribution. However, in the later stage, neural networks may overfit noisy labels, causing noisy-label samples to deviate from the original class distribution and eventually leading to incorrect classification.

	When designing robust loss functions, a common assumption is that samples within the same class are drawn from an identical distribution. Extending this assumption to the overfitting scenario, we further hypothesize that samples of the same class associated with distinct noisy labels also tend to exhibit similar overfitted feature distributions.
	That is, when a neural network overfits a particular noise-related feature, this feature is not confined to a single noisy-label category. 
	As illustrated in the right panel of Fig.~\ref{fig_mov}, samples from true class 1 are mislabeled as classes 3 and 4, and these samples may share the same overfitted features. In this case, an increase in the probability of noisy class 3 may be accompanied by a substantial increase in the probability of class 4. This motivates us to leverage relationships among class probabilities in the loss design to alleviate overfitting to noisy labels.
	
	As discussed above, to preserve inter-class separation and maintain training stability, we design two families of losses under inter-class independent and dependent formulations, referred to as Target Separation and Binary Reduction, respectively. We further investigate their theoretical properties, including symmetry and asymmetry conditions for noise robustness, and establish corresponding sufficient conditions under different formulations. Based on the proposed framework, we construct two robust loss instances and validate their effectiveness on both synthetic and real-world noisy-label benchmarks. Our contributions are summarized as follows:
	
	\begin{itemize}
		\item We propose a new framework for robust loss construction based on univariate base functions, providing a flexible and general formulation for noisy-label learning.
		
		\item We develop two general formulations, namely target separation and binary reduction, and theoretically investigate their properties, including symmetry and asymmetry conditions for noise robustness.
		
		\item Based on the proposed framework, we construct two robust loss instances and validate their effectiveness through extensive experiments on synthetic and real-world noisy-label benchmarks.
	\end{itemize}
	
	Overall, this work provides a systematic perspective on robust loss design for noisy-label learning. Instead of treating robust losses as isolated objectives, we formulate them as induced losses generated from simple univariate base functions. This perspective allows different robustness properties to be analyzed and controlled through the structure of the base function and the corresponding multiclass construction. The remainder of this paper is organized as follows. We first review related work on noisy-label learning and robust loss functions. We then present the proposed loss construction framework and analyze the theoretical properties of Target Separation and Binary Reduction. Finally, we instantiate the framework with concrete loss functions and evaluate them on synthetic and real-world noisy-label benchmarks.

	\section{Related Work}
	Existing approaches for learning with noisy labels can be broadly categorized into four groups: improved training strategies~\cite{Anurag2019SeCoST, Eran2017Decoupling}, label correction~\cite{Tong2015Clothing1M, Lee2018CleanNet}, loss correction~\cite{Szegedy2016Rethinking, Han2018Masking}, and robust losses~\cite{Zhang2018GCE, Wei2023CELC}. 
	Improved training strategies design adaptive training schemes, such as mentor--student frameworks~\cite{Anurag2019SeCoST} or co-training paradigms~\cite{Eran2017Decoupling}, to mitigate the impact of noisy supervision. 
	Label correction methods aim to improve label quality by identifying and correcting corrupted labels, often relying on auxiliary clean data or additional models such as graphical models~\cite{Tong2015Clothing1M} and neural networks~\cite{Lee2018CleanNet, Veit2017Learninig}. 
	Loss correction methods enhance robustness by modifying the loss function, for example through Label Smoothing Regularization (LSR)~\cite{Szegedy2016Rethinking} or estimating noise transition matrices~\cite{Han2018Masking}; however, their effectiveness depends heavily on accurate noise estimation. 
	These methods typically require complex training procedures, additional networks, or careful hyperparameter tuning.
	
	In contrast, robust losses provide a simpler and more general solution by directly designing loss functions that are inherently insensitive to label noise. Early studies~\cite{Aritra2017Robust, Ma2020NCERCE, Naresh2013Tolerance, van2015Learning} have shown that loss functions satisfying the symmetry condition are inherently robust to label noise. 
	However, such symmetric losses are often difficult to optimize due to their overly strict constraints, as exemplified by Mean Absolute Error (MAE). 
	To address this issue, subsequent work has proposed a variety of hybrid loss functions that combine the robustness with the strong fitting ability, including Generalized Cross Entropy (GCE)~\cite{Zhang2018GCE}, Symmetric Cross Entropy (SCE)~\cite{Wang2019SCE}, Taylor Cross Entropy~\cite{Lei2020Taylor}, and Jensen--Shannon divergence-based loss~\cite{Erik2021JS}.

	Beyond mixture designs, other approaches relax the symmetry constraint~\cite{Zhou2021SR, Wang2024epsilon} or exploit complementary loss structures~\cite{Kim2019NLNL, Kim2021JNPL}. 
	For instance, methods such as Sparse Regularization~\cite{Zhou2021SR} and $\epsilon$-Softmax~\cite{Wang2024epsilon} approximate one-hot labels to achieve relaxed symmetry, while approaches like Active Passive Loss (APL)~\cite{Ma2020NCERCE} and Active Negative Loss (ANL)~\cite{Ye2023ANL} jointly employ multiple loss components to improve fitting ability. 
	In addition, techniques such as Negative Label Smoothing~\cite{Wei2022To}, PHuber-CE\cite{Menon2020PHuber} and LogitClip~\cite{Wei2023CELC} further enhance robustness by mitigating overfitting to noisy labels through label, gradient or logit regularization. 
	Recently, asymmetric loss functions~\cite{Zhou2021Asymmetric, Zhou2023Asymmetric} have been proposed to better handle practical noise settings, demonstrating improved empirical performance over symmetric losses. Furthermore, passive asymmetric loss~\cite{Wang2025JAL} and Variation-Bounded Loss~\cite{wang2026variation} have been explored.
	
	Our framework is developed based on symmetric losses~\cite{Aritra2017Robust, Charoenphakdee2019On, Ma2020NCERCE} and the asymmetric loss formulation introduced by Zhou et al~\cite{Zhou2021Asymmetric, Zhou2023Asymmetric}. 
	It simplify complex robust objective design into combinations of univariate functions, thereby providing a new perspective for robust loss construction.

	\section{Preliminaries}
	
	\subsection{Problem Definition}
	We consider a classification problem with $K$ classes, input space $\mathcal{X} \subset \mathbb{R}^d $ and label space $\mathcal{Y} = [K] = \{1, 2, \dots, K\}$. We focus on single-label classification, where each sample $\mathbf{x} \in \mathcal{X}$ has exactly one correct label $y \in \mathcal{Y}$, such that $p(y | \mathbf{x}) = 1$ and $p(k | \mathbf{x}) = 0$ for all $k \neq y$. In the noise-free setting, the training dataset is denoted by $\mathcal{D} = \{(x_i, y_i)\}_{i=1}^N$. A classifier $f : \mathcal{X} \to \mathcal{U}$ maps each input $\mathbf{x} \in \mathcal{X}$ to a probability vector $\mathbf{u} \in \mathcal{U}$, where $\mathcal{U} = \{ \mathbf{u} \in [0,1]^K \mid \sum_{i=1}^{K} u_i = 1 \}$ is the probability simplex. The predicted class is given by $k = \arg\max_{i \in [K]} u_i$. The loss function $L : \mathcal{U} \times \mathcal{Y} \rightarrow \mathbb{R}$ measures the discrepancy between the predicted probability vector $\mathbf{u}$ and the true label $y$.
	
	Under label noise, the label $y$ may be flipped, resulting in a noisy training dataset $\tilde{\mathcal{D}} = \{(x_i, \tilde{y}_i)\}_{i=1}^N$. The label corruption process can be formulated as follows:
	\begin{equation}
		\tilde{y} = 
		\begin{cases} 
			y & \text{with probability } \eta_{\mathbf{x}, y} = 1 - \tilde{\eta}_{\mathbf{x}}, \\
			k, k \in [K]\setminus \{y\}    & \text{with probability } \eta_{\mathbf{x}, k},
		\end{cases} \nonumber
	\end{equation}
	where $\tilde{\eta}_{\mathbf{x}} = \sum_{k \neq y} \eta_{\mathbf{x}, k}$, and $\eta_{\mathbf{x}, k}$ denotes the probability that the true label $y$ is flipped to class $k$, i.e., $\eta_{\mathbf{x}, k} = P(\tilde{y} = k \mid y, \mathbf{x})$.
	
	To characterize different noise behaviors, we consider three widely-used label noise models following~\cite{Xia2020Part, Ye2023ANL}. Under symmetric noise, a true label may flip uniformly to any other class. The transition probabilities satisfy $\eta_{\mathbf{x},y}=1-\eta$ and $\eta_{\mathbf{x},k}=\frac{\eta}{K-1}$ for $k\neq y$, where $\eta$ is a global noise rate shared by all samples. Under asymmetric noise, the corruption pattern depends on the class of the true label, and the flipping probability is governed by class-dependent noise rates $\eta_y$. For instance-dependent noise, the corruption probability varies across individual samples, where the probability of label corruption is controlled by an instance-specific parameter $\eta_{\mathbf{x}}$. Symmetric noise corresponds to uniform label corruption, while asymmetric and instance-dependent noise represent more realistic noise scenarios.

	\subsection{Noise Risk}
	Given a loss function $L$, the $L$-risk~\cite{Bartlett2006Risk} of a classifier $f$ on the clean dataset $\mathcal{D}$ is defined as
	\[
	R_L(f) = \mathbb{E}_{\mathcal{D}}[L(f(\mathbf{x}),y)] = \mathbb{E}_{(\mathbf{x},y)}[L(f(\mathbf{x}),y)].
	\]
	The goal of neural network training is to minimize the $L$-risk and obtain the optimal classifier $f^* \in \arg\min_f R_L(f)$. Under label noise, the risk becomes $ \tilde{R}_L(f) = \mathbb{E}_{(\mathbf{x},\tilde{y})}[L(f(\mathbf{x}),\tilde{y})] $. Accordingly, the optimal classifier is defined as $f_\eta^* \in \arg\min_f \tilde{R}_L(f)$. Throughout this paper, we assume that both $f^*$ and $f_\eta^*$ are unique. We follow the definition of noise tolerance introduced in~\cite{Naresh2013Tolerance}.
	
	\begin{definition}[Noise Tolerance] \label{def_1}
		Let $f^*$ and $f_\eta^*$ be the optimal classifiers minimizing the clean risk and noisy risk under loss $L$, respectively. 
		The risk minimization under $L$ is said to be noise-tolerant if
		\[
		\mathbf{P}_{\mathcal{D}}[ \mathrm{pred} \circ f^*(\mathbf{x}) = y_\mathbf{x}] 
		=
		\mathbf{P}_{\mathcal{D}}[\mathrm{pred} \circ f_\eta^*(\mathbf{x}) = y_x],
		\]
		where $\mathrm{pred} \circ f^*(\mathbf{x})$ denotes the predicted class.
	\end{definition}
	
	Clearly, if $\mathrm{pred} \circ f^*(\mathbf{x}) = \mathrm{pred} \circ f_\eta^*(\mathbf{x})$, then the loss $L$ is noise-tolerant. In many existing works~\cite{Zhou2021Asymmetric, Zhou2023Asymmetric, Wang2025JAL}, a stronger condition $f^* = f_\eta^*$ has been studied. This condition implies that minimizing the risk yields the same classifier under both noisy and noise-free settings.

	\subsection{Symmetric and Asymmetric Losses}
	One of the most widely studied families of noise-robust loss functions is the class of symmetric losses~\cite{van2015Learning, Aritra2017Robust, Charoenphakdee2019On}, which are defined as follows.
	\begin{definition}[Symmetric Loss]
		A loss function $L$ is said to be symmetric if it satisfies
		\[
		\sum_{i=1}^{K} L(\mathbf{u}, i) = C,
		\]
		for all prediction vectors $\mathbf{u}$, where $C$ is a constant.
	\end{definition}
	
	When clean labels dominate the dataset, symmetric loss functions have been shown to be noise-tolerant. Recently, Zhou et al.~\cite{Zhou2021Asymmetric, Zhou2023Asymmetric} proposed a broader class of robust loss functions, referred to as asymmetric losses.

	\begin{definition}[Asymmetric Loss]
		Given non-negative weights $w_1,\dots,w_K$, suppose there exists 
		$t \in [K]$ such that $w_t > \max_{i \neq t} w_i$. 
		A loss function $L$ is said to be asymmetric if
		\[
		\arg\min_{\mathbf{u}} \sum_{i=1}^K w_i L(\mathbf{u}, i)
		=
		\arg\min_{\mathbf{u}} L(\mathbf{u}, t),
		\]
		where $\mathbf{u}$ denotes the prediction score vector and the minimizer is the one-hot vector $\mathbf{e}_t$.
	\end{definition}
	
	Symmetric loss functions are a special case of asymmetric loss functions. Furthermore, when clean labels dominate, asymmetric loss functions are also noise-tolerant.
	
	\begin{figure*}[htbp]
		\centering
		\includegraphics[width=0.99\linewidth]{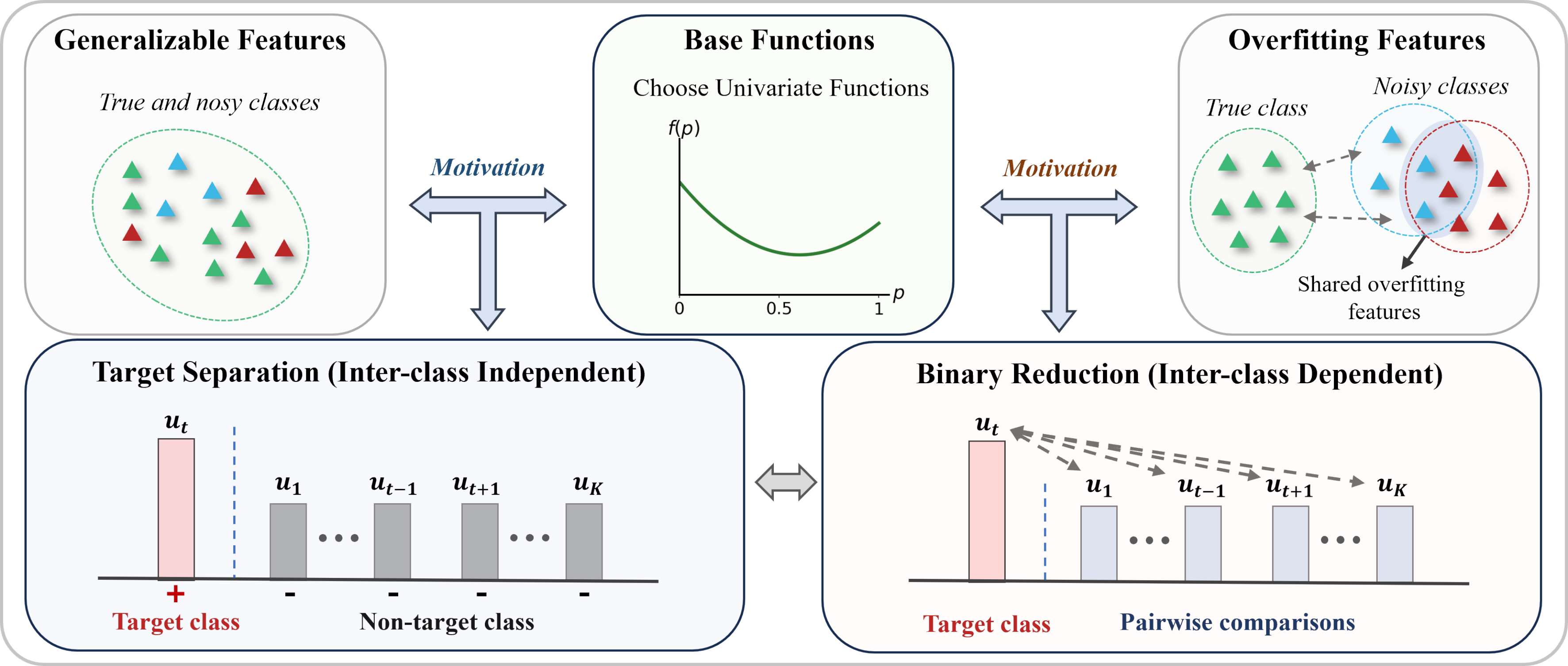}
		\caption{
			Overview of the proposed framework. Starting from univariate base functions, we derive two complementary robust loss formulations. Target Separation is motivated by generalizable class-level features and treats classes independently by promoting the target class while suppressing non-target classes. Binary Reduction is motivated by shared overfitting features among noisy classes and captures inter-class dependencies through pairwise comparisons between the target class and each non-target class.
		}
		\label{fig_overview}
	\end{figure*}

	\section{Methodology}

	\begin{figure}[htbp]
		\centering
		
		\subfloat{
			\includegraphics[width=0.49\linewidth]{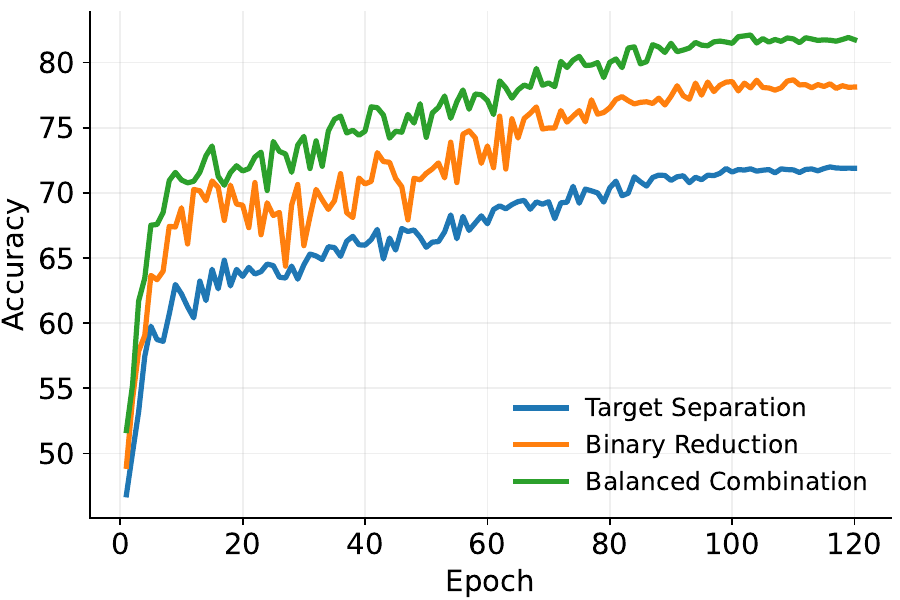}
		}
		\subfloat{
			\includegraphics[width=0.49\linewidth]{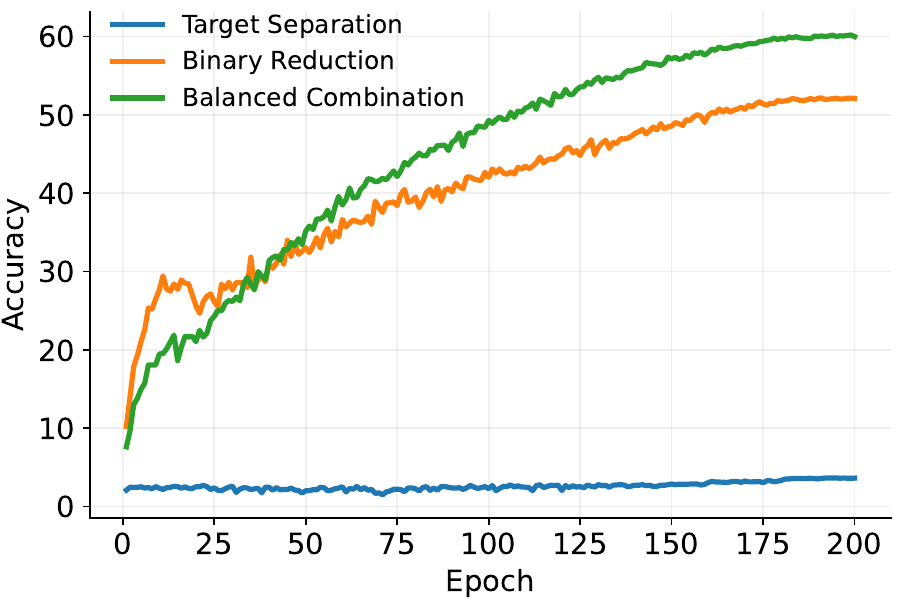}
		}

		\caption{Training curves of the target-separation term only, the binary-reduction term only, and their balanced combination on CIFAR-10N (left) and CIFAR-100N (right). While each individual component captures only part of the desired robustness, their balanced combination consistently achieves higher performance, indicating a clear synergistic effect between target separation and binary reduction.}
		\label{fig_balance}
	\end{figure}

	\subsection{Method Overview}
	
	We now describe the proposed framework for constructing robust multiclass losses from univariate base functions. Let $\mathbf{u}=(u_1,\ldots,u_K)^T$ denote the predicted class-probability vector, and let $t\in\{1,\ldots,K\}$ denote the observed label, which may be corrupted in noisy-label learning. Instead of designing a multiclass loss $\mathcal{L}(\mathbf{u},t)$ directly, we first specify a univariate base function and then induce a multiclass loss through a structured mapping. In this way, the shape of the base function controls the local response of the loss to class probabilities, while the mapping determines how the target and non-target classes interact.
	
	Fig.~\ref{fig_overview} illustrates the overall framework. We consider two complementary mappings from base functions to multiclass losses. The first mapping, Target Separation, constructs the loss by combining a target-class term with an aggregated non-target term. It treats different class probabilities in an inter-class independent manner and explicitly balances target promotion and non-target suppression. This formulation is motivated by the observation that, when clean labels dominate, samples from the same class tend to share generalizable class-level features. In this case, preserving class-wise separation and preventing noisy labels from dominating the optimization are essential for robust training.
	
	The second mapping, Binary Reduction, constructs the loss through pairwise comparisons between the observed class and each non-target class. Unlike Target Separation, this formulation does not evaluate each class probability independently. Instead, it compares the relative probabilities within each target--non-target pair, thereby introducing inter-class dependencies into the loss. This is motivated by the overfitting behavior discussed in the introduction: noise-related features learned during overfitting may be shared by multiple classes, causing the probabilities of irrelevant classes to increase simultaneously. Modeling such pairwise relationships provides an additional mechanism to alleviate overfitting to noisy labels.

	The complementary roles of the two mappings are also reflected empirically. As shown in Fig.~\ref{fig_balance}, using either Target Separation or Binary Reduction alone yields lower accuracy, whereas their balanced combination consistently performs better on both CIFAR-10N and CIFAR-100N.  These results suggest that the two formulations are not redundant, but instead provide complementary mechanisms for improving robustness under human-annotated label noise.
	
	Under the two mappings, robustness-related properties of the induced losses can be analyzed through simple properties of the corresponding base functions. For Target Separation, the key factors are the separation coefficient and the derivative range of the univariate base function. For Binary Reduction, the analysis relies on the symmetry and derivative behavior of the underlying binary loss when evaluated on pairwise conditional probabilities. The following subsections present the two formulations in detail, analyze their symmetry and asymmetry properties, and then instantiate concrete robust losses under the derived theoretical conditions.

	\subsection{Target Separation}
	Through normalization, a general loss function can be transformed into a symmetric loss. However, empirical studies suggest that such normalized losses are often difficult to optimize in practice. Although asymmetric losses provide more general conditions for noise robustness, designing practically effective asymmetric losses remains challenging. Therefore, we propose the idea of constructing desired loss functions from univariate base functions. 
	
	The first goal is to ensure that, when clean labels dominate and samples from the same class share a common feature distribution, clean-label samples can guide the distribution toward the correct optimization direction. In addition to losses such as cross-entropy that directly optimize the labeled class, Negative Learning~\cite{Kim2019NLNL} has also been shown to be effective for classification by minimizing the probabilities assigned to non-label classes. Inspired by these two types of losses, we propose target-separated losses that unify them within a single framework and establish sufficient conditions for their noise robustness. It is defined as follows.

	\begin{definition}[Target-Separated Loss]
		\label{de_TSL}
		A loss function $\mathcal{L}$ is said to be target-separated if it can be written in the following form.
		\[
		\mathcal{L}(\mathbf{u}, t) = L(\mathbf{u}, t) - \frac{\tau}{K-1} \sum_{i \neq t}  L(\mathbf{u}, i).
		\]
		Here $K$ denotes the number of classes, $\tau \ge 0$ is a constant, and we let $a = \frac{\tau}{K-1}$ for notational convenience.
	\end{definition}

	The target-separated loss modifies the base loss by introducing an additional term over non-target classes. The coefficient $a$ controls the relative strength of this non-target component. This form is related in spirit to complementary-label and negative-learning-based losses, such as NLNL~\cite{Kim2019NLNL} and RCE~\cite{Wang2019SCE}, since these methods also exploit information from labels other than the observed target. However, in the proposed formulation, both the target and non-target terms are induced from the same base loss $L(\mathbf{u}, i)$. This shared structure allows us to analyze the robustness of the loss through the coefficient $a$ and the properties of the base function.
	
	We first consider the symmetry property of the target-separated loss. Since the non-target term appears for every label, the total loss over all possible labels depends on the balance between the target term and the aggregated non-target terms. The following theorem shows that a specific choice of $a$ leads to a symmetric loss. The proofs of all theoretical results are provided in the Appendix.

	\begin{theorem}
		\label{thm_1}
		If $a = \frac{1}{K-1}$, then the target-separated loss $\mathcal{L}$ is symmetric.
	\end{theorem}

	Theorem~\ref{thm_1} shows that symmetry can be achieved by choosing $a=1/(K-1)$. This result does not rely on a particular choice of the base loss $L$, and therefore characterizes a general property of the target-separated form. However, symmetry only describes the behavior of the loss after summing over all labels. To further understand how the loss behaves under a weighted label distribution, we next study its response to a local change in the prediction vector.
	
	Specifically, we consider the case where $L(\mathbf{u}, i)=h(u_i)$. Given a set of label weights $(w_1,\ldots,w_K)$, we analyze whether transferring a small amount of prediction mass from a class $s$ to a class $t$ can reduce the weighted target-separated loss. This local comparison is useful because it directly reflects how the loss encourages the prediction vector to move toward classes with larger weights. The following lemma shows that this finite-shift condition is equivalent to a derivative condition on the base function $h$.
	﻿

	\begin{lemma}
		\label{lem_1}
		Let $h(x)$ be a differentiable and monotonically decreasing function on $[0,1]$ and let $L(\mathbf{u}, i) = h(u_i)$. Given weights $w_1, \ldots, w_K$ with $w_t > w_i, \forall i\neq t$. For all $s \neq t$, the following two conditions are equivalent. 
		
		(1) For all $0 \leq u_s, u_t \leq 1$ and $0 \le \triangle u \le u_s$, the following inequality holds.
		\[
		\sum_{i=1}^K w_i \mathcal{L}(\mathbf{u}, i) \geq \sum_{i=1}^K w_i \mathcal{L}\left(\mathbf{u} + \triangle \mathbf{u}_t  - \triangle \mathbf{u}_s, i\right),
		\]
		where $\triangle \mathbf{u}_t = \triangle u * \mathbf{e}_t$, and $\triangle \mathbf{u}_s$ is defined analogously.
		
		(2) For all $0 \leq u_s, u_t \leq 1$, the following inequality holds.
		\[
		\left(  (1- w_t) a -w_t \right) h^{\prime}  (u_t) \geq \left( (1-w_s) a - w_s  \right)   h^{\prime} (u_s).
		\]
		
	\end{lemma}

	Lemma~\ref{lem_1} provides a bridge between the target-separated construction and its local optimization behavior. Condition (1) describes the desired change of the weighted loss under a probability shift from class $s$ to class $t$, while condition (2) expresses the same requirement using the derivative of the base function. Therefore, instead of verifying the weighted loss inequality for all possible shifts, it is sufficient to check a pointwise inequality involving $a$, the label weights, and $h^{\prime}$.
	
	This derivative characterization leads directly to a sufficient condition for asymmetry. If the inequality in Lemma~\ref{lem_1} holds for every non-target class $s$, then increasing the prediction assigned to the larger-weight class $t$ does not increase the weighted target-separated loss. This gives the following theorem.
	
	\begin{theorem}
		\label{thm_2}
		Let $h(x)$ be a differentiable and monotonically decreasing function on $[0,1]$ and let $L(\mathbf{u}, i) = h(u_i)$. Given weights $w_1, \ldots, w_K$ with $w_t > w_i, \forall i\neq t$. Then, $\mathcal{L}$ is asymmetric if  $\forall 0 \leq u_s, u_t \leq 1$,
		\[
		\left(  (1- w_t) a -w_t \right) h^{\prime}  (u_t) \geq \left( (1-w_s) a - w_s  \right) h^{\prime} (u_s).
		\]
	\end{theorem}
	
	Theorem~\ref{thm_2} shows that the asymmetry of the target-separated loss depends jointly on the separation coefficient $a$ and the derivative profile of the base function $h$. The coefficient $a$ determines how strongly the non-target losses affect the weighted loss, whereas $h^{\prime}$ determines the local sensitivity of the base loss to changes in the prediction probability. Thus, the theorem provides a direct design criterion for constructing asymmetric target-separated losses.
	
	Although the condition in Theorem~\ref{thm_2} is compact, it does not fully characterize the symmetric case, which can also be regarded as a special case under the asymmetric formulation. To make the condition easier to verify and interpret, we further derive several sufficient regimes according to the relative magnitude of $a$. These regimes are summarized in the following corollary.

	\begin{corollary}
		\label{cor_1}
		Under the conditions of Theorem \ref{thm_2} and $0 < w_t < 1$, $\mathcal{L}$ is asymmetric whenever any of the following conditions holds.
		
		(1)  
		\[
		\max_{i\neq t} \frac{w_i}{1 - w_i} \leq a \leq \frac{w_t}{1 - w_t}.
		\]
		
		(2) When $a < \max_{i\neq t} \frac{w_i}{1 - w_i} $, 
		\[
		\frac{h'(u_t)}{h' (u_s)} \geq \frac{(1-w_s)a - w_s}{(1-w_t)a - w_t}, \quad \forall s \neq t.
		\]
		
		(3) When $a > \frac{w_t}{1 - w_t}$, 
		\[
		\frac{h'(u_t)}{h' (u_s)} \leq \frac{(1-w_s)a - w_s}{(1-w_t)a - w_t}, \quad \forall s \neq t.
		\]
		
		(4)
		\[
		a = \frac{1}{K - 1}.
		\]
		
	\end{corollary}

	Corollary~\ref{cor_1} gives more explicit conditions under which the derivative inequality in Theorem~\ref{thm_2} holds. When $a$ lies between the non-target and target weight ratios, the coefficient itself is sufficient to ensure the desired inequality. When $a$ is outside this interval, the derivative ratio of the base function needs to compensate for the imbalance introduced by $a$. Therefore, the robustness behavior of target-separated losses is controlled by two factors: the separation coefficient $a$ and the range of the derivative $h^{\prime}$.
	
	These results suggest that the proposed target-separated formulation provides a unified way to construct robust losses from a base function. The coefficient $a$ controls the contribution of non-target classes, while the base function determines how gradients are assigned to different prediction probabilities. This is consistent with the intuition that noisy labels should not dominate the optimization through disproportionately large gradients. Notably, the target-separated loss reduces to an active loss~\cite{Ma2020NCERCE} when ($a=0$). In addition, some losses, such as AUL, AGCE, and AEL~\cite{Zhou2021Asymmetric}, can be seen as special cases under Theorem~\ref{thm_2}.
	﻿

	When clean labels dominate the training distribution, these properties can lead to a noise-tolerant solution at the expected-risk level. As illustrated in the middle of Fig.~\ref{fig_mov}, this can be interpreted as clean-label samples guiding the optimization of noisy-label samples. However, late-stage overfitting to noisy labels may still occur. Complex noise patterns can drive noisy samples in different directions, increasing the probabilities of classes unrelated to both the true and noisy labels. Even under a single noise transition, moving noisy samples from the true-class distribution toward the corresponding noisy-class distribution may increase the probabilities of other irrelevant classes. We therefore then exploit inter-class dependencies so that these irrelevant classes act as gatekeepers against such distributional drift.
	﻿

	\subsection{Binary Reduction}
	While target-separated losses enjoy desirable theoretical properties, the separation among variables prevents them from explicitly modeling inter-class dependencies. To address this limitation, we turn to binary classification, where the inter-class relationship admits a natural univariate characterization. Let $l(p,q)$ be an arbitrary binary loss defined on the binary simplex, where $p+q=1$. Since $q=1-p$, the binary objective can be equivalently represented as $l(p,1-p)$. This univariate representation does not decouple the two classes; rather, the two probabilities remain inherently coupled through the shared simplex constraint. Therefore, a univariate objective can still characterize the inter-class dependency in binary classification.
	Motivated by this observation, we propose a binary reduction approach that decomposes the multiclass objective into multiple binary objectives, thereby introducing inter-class dependencies through pairwise class comparisons.

	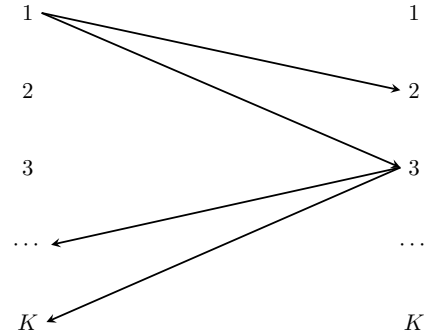
\begin{figure}[htbp]
		\centering
		\resizebox{0.65\linewidth}{!}{
			\begin{tikzpicture}[>=stealth, thick]
				
				\node (L1)  at (0,  0) {$1$};
				\node (L2)  at (0, -1.2) {$2$};
				\node (L3)  at (0, -2.4) {$3$};
				\node (Ld)  at (0, -3.6) {$\cdots$};
				\node (L10) at (0, -4.8) {$K$};

				\node (R1)  at (6,  0) {$1$};
				\node (R2)  at (6, -1.2) {$2$};
				\node (R3)  at (6, -2.4) {$3$};
				\node (Rd)  at (6, -3.6) {$\cdots$};
				\node (R10) at (6, -4.8) {$K$};

				\draw[->] (L1.east) -- (R2.west);
				\draw[->] (L1.east) -- (R3.west);
				\draw[->] (R3.west) -- (Ld.east);
				\draw[->] (R3.west) -- (L10.east);
		\end{tikzpicture}}
		\caption{$K$-class classification via (K-1) binary comparisons.}
		\label{fig_1}
	\end{figure}

	We still consider a multiclass classification problem with $K$ classes. And the probability vector output by the network model is denoted by $\mathbf{u} = (u_1, u_2, \dots, u_K)^T$. Under the Bayesian framework, the classification between classes $i$ and $j$ can be viewed as a conditional classification problem given that the sampled data belongs to $\{i,j\}$. The corresponding conditional probabilities are
	\[
	p(i \mid i, j) = \frac{u_i}{u_i + u_j} , \quad p(j \mid i, j) = \frac{u_j}{u_i + u_j}.
	\]

	Fig.~\ref{fig_1} illustrates how $K$-class classification is implemented through binary classification. Starting from the first two classes, we retain the class with the larger conditional probability and compare it sequentially with each remaining class. Since, for any $i\neq t$,
	\[
	u_t \geq u_i \quad \Longleftrightarrow \quad p(t\mid i,t)\geq p(i\mid i,t),
	\]
	the class returned by this procedure is exactly the class with the largest multiclass probability.

	A multiclass prediction can be viewed through pairwise discrimination between classes. Let $l(p,1-p)$ denote a binary classification loss, where $p$ is the probability assigned to one class after normalization within a given class pair. For a sample with target label $t$, multiclass learning requires distinguishing $t$ from every non-target class $i\neq t$. This motivates the following binary-reduced construction, which aggregates the pairwise losses between the target and each non-target class.
	﻿

	\begin{definition}[Binary-Reduced Loss]
		\label{de_BRL}
		A loss function $\mathcal{L}$ is said to be binary-reduced if it can be written in the following form.
		\[
		\mathcal{L} (\mathbf{u}, t) = \frac{1}{K-1} \sum_{i \neq t} \ell(u_t, u_i),
		\]
		where
		\begin{equation}
			\ell(u_t, u_i) =
			\begin{cases} 
				l\left(\frac{u_t}{u_t+u_i}, \frac{u_i}{u_t+u_i}\right), & u_t + u_i \neq 0, \\
				\xi , & u_t + u_i = 0.
			\end{cases}
			\nonumber
		\end{equation}
		Here $\xi$ is a constant and $l$ is a binary classification loss.
	\end{definition}

	The case $u_t+u_i=0$ is defined separately to avoid ambiguity when both probabilities vanish. For all non-degenerate pairs, the binary loss $l$ is applied  to the conditional probabilities. Therefore, each term in the binary-reduced loss depends on the relative preference between the observed class and one non-target class, rather than on their absolute probability values.
	
	Although $l$ is written as a bivariate function, its inputs after normalization always lie on the binary simplex. Hence, the behavior of $l$ in the binary-reduced loss can be characterized by the univariate restriction $l(p,1-p)$. This observation allows us to analyze the multiclass loss through a binary loss while retaining the pairwise structure among classes. We first study the symmetry property and examine whether the symmetry of the binary loss is preserved after binary reduction.

	\begin{theorem}
		\label{thm_3}
		Suppose that $\xi = l(0.5, 0.5)$. The binary loss $l$ is symmetric if and only if the binary-reduced loss $\mathcal{L}$ is symmetric.
	\end{theorem}

	Theorem~\ref{thm_3} shows that the binary-reduction operation preserves symmetry when the degenerate value is chosen consistently as $\xi=l(0.5,0.5)$. In other words, under this choice of $\xi$, the multiclass binary-reduced loss is symmetric exactly when the underlying binary loss is symmetric. Therefore, the symmetry analysis of $\mathcal{L}$ can be reduced to the symmetry analysis of $l$ on the binary simplex.
	
	To obtain a more explicit condition, we introduce a centered univariate function $g(p)=l(p,1-p)-l(0.5,0.5)$. This centering removes the constant value at the midpoint and makes the symmetry condition depend only on the relation between $p$ and $1-p$.
	
	\begin{theorem}
		\label{thm_4}
		Let $g(p) = l(p,1-p) - l(0.5,0.5)$. Then $l$ is symmetric if and only if $g(p)$ is odd about $p=0.5$.
	\end{theorem}
	
	Theorem~\ref{thm_4} gives a simple characterization of binary symmetry: after centering at $p=0.5$, the function $g$ must be odd about the midpoint. Combining this result with Theorem~\ref{thm_3}, we can directly obtain the corresponding symmetry condition for the binary-reduced loss.
	
	\begin{corollary}
		Suppose that $\xi = l(0.5, 0.5)$. The binary-reduced loss $\mathcal{L}$ is symmetric if and only if $g(p) = l(p,1-p) - l(0.5,0.5)$ is odd about $p=0.5$.
	\end{corollary}
	
	The corollary shows that the symmetry of the multiclass binary-reduced loss can be verified entirely through the centered univariate function $g$. This result is useful for loss construction because it avoids analyzing the multiclass loss directly: once the binary loss is centered at $p=0.5$, symmetry reduces to the simple condition $g(p)+g(1-p)=0$. We next turn to the asymmetric case, where the goal is to characterize how the weighted binary-reduced loss changes under probability shifts toward classes with larger weights.
	Since $\mathcal{L}$ is constructed as an average of pairwise binary losses, it is natural to first impose a monotonicity condition on each binary comparison.
	
	Specifically, for two classes with weights $w_i \geq w_j$, we require that transferring probability mass from the lower-weight side to the higher-weight side should not increase the weighted binary loss. The following lemma shows that this pairwise condition is sufficient to ensure the desired multiclass behavior of the binary-reduced loss.

	\begin{lemma}
		\label{lem_2}
		Let $ \mathcal{L} (\mathbf{u}, i) = \frac{1}{K-1} \sum_{j \neq i} \ell\left( u_i, u_j \right) $. Suppose that $\xi = l(1,0) = \inf l(p, q)$ and $l(0,1) = \sup l(p, q)$. Without loss of generality, assume that 
		$w_1 > w_2 \geq w_3 \geq \cdots \geq w_K$. If for any $w_i \geq w_j$ with $i \neq j$, if $0 \le p, q \le 1$ and $0 \le \triangle p \le q$, the following inequality holds,
		\[
		w_i l(p, q) + w_j l(q,p) \geq w_i l(p + \triangle p , q - \triangle p ) + w_j l(q - \triangle p , p + \triangle p ),
		\]
		then, for all $1 < s \leq K$, the following inequality holds.
		\[
		\sum_{i=1}^K w_i \mathcal{L}(\mathbf{u}, i) \geq \sum_{i=1}^K w_i \mathcal{L}\left(\mathbf{u} + \sum_{j=s}^{K} ( u_j * \mathbf{e}_1  - u_j * \mathbf{e}_j ), i\right).
		\]
		This implies that $\mathcal{L}$ is asymmetric.
		
	\end{lemma}
	
	Lemma~\ref{lem_2} connects the pairwise behavior of the binary loss $l$ to the global behavior of the multiclass loss $\mathcal{L}$. The assumption in the lemma is imposed only on a two-class comparison, but the conclusion applies to the binary-reduced loss after aggregating all pairwise terms. Thus, asymmetry of $\mathcal{L}$ can be established by verifying a pairwise transfer condition for $l$.
	
	However, the condition in Lemma~\ref{lem_2} still involves all possible transfer magnitudes $\triangle p$. To make the condition easier to verify and more useful for loss design, we next rewrite it as a derivative condition using the centered function $g$.
	
	\begin{lemma}
		\label{lem_3}
		Let $g(p) = l(p,1-p) - l(0.5,0.5)$. Suppose that $g(p)$ is differentiable on $[0,1]$. Given any weights $w_1, w_2$ with $w_1 \geq w_2$, the following two conditions are equivalent.
		
		(1) For all $0 \leq p, q \leq 1$ and $0 \le \triangle p \le q$, 
		\[
		w_1 l(p, q) + w_2 l(q,p) \geq w_1 l(p + \triangle p , q - \triangle p ) + w_2 l(q - \triangle p , p + \triangle p ).
		\]
		
		(2) For all $0 \leq p \leq 1$,
		\[
		w_1 g'(p) \leq w_2 g'(1-p).
		\]
		
	\end{lemma}
	
	Lemma~\ref{lem_3} shows that the finite pairwise transfer condition is equivalent to a pointwise inequality involving $g'$. This equivalence is important because it replaces a condition over all possible probability shifts with a local condition on the derivative of the centered binary loss. Consequently, the asymmetry of the binary-reduced loss can be analyzed through the relative values of $g'(p)$ and $g'(1-p)$.
	
	By combining Lemma~\ref{lem_2} and Lemma~\ref{lem_3}, we obtain a sufficient condition for the asymmetry of binary-reduced losses. The resulting condition directly links the robustness of $\mathcal{L}$ to the derivative profile of the base function.
	
	\begin{theorem}
		\label{thm_5}
		Let $g(p) = l(p,1-p) - l(0.5,0.5)$ and $ \mathcal{L} (\mathbf{u}, i) = \frac{1}{K-1} \sum_{j \neq i} \ell\left( u_i, u_j \right) $. Suppose that $\xi = l(1,0) = \inf l(p, q)$, $l(0,1) = \sup l(p, q)$ and $g(p)$ is differentiable on $[0,1]$. Given any weights $w_1, \ldots, w_K$ with $w_t > w_i, \forall i\neq t$, $\mathcal{L}$ is asymmetric, if  for all $w_i \geq w_j, i\neq j$,  $0 \leq p \leq 1$,
		\[
		w_i g'(p) \leq w_j g'(1-p).
		\]
		Equivalently, for all $p \in [0,1]$, $g'(p) \leq 0$ and one of the following cases holds.
		
		(1) If $g'(p) = 0$, $g'(1-p) = 0$.
		
		(2) For all $w_i \geq w_j, i\neq j$, and $w_j >0$, if $g'(p) < 0$,  
		$
		\frac{g'(1-p)}{g'(p)} \leq \min \left\{\frac{w_i}{w_j}\right\} .
		$
		
	\end{theorem}
	
	Theorem~\ref{thm_5} shows that the asymmetry of the binary-reduced loss is governed by the derivative relation between $g'(p)$ and $g'(1-p)$. The condition $w_i g'(p) \leq w_j g'(1-p)$ ensures that, for each pairwise comparison, the weighted loss does not increase when probability mass is shifted toward the class with the larger weight. The equivalent cases further indicate that $g'$ should be non-positive, and when $g'(p)$ is strictly negative, the ratio between $g'(1-p)$ and $g'(p)$ must be controlled by the corresponding weight ratios.
	
	Overall, these results use binary classification losses as an intermediate bridge between univariate base functions and multiclass binary-reduced losses. The symmetry of the binary-reduced loss is characterized by the oddness of the centered function $g$ about $p=0.5$, while its asymmetry is controlled by the weighted derivative relation between $g'(p)$ and $g'(1-p)$. By decomposing multiclass learning into multiple normalized pairwise comparisons, the binary-reduced framework explicitly captures inter-class dependencies and therefore complements Target Separation.

	The pairwise conditional probability also exhibits a useful sensitivity property. Since $p(t \mid i,t) =\frac{u_t}{u_i+u_t}$, the probability becomes highly sensitive to small changes in $u_i$ around zero when $t$ is a noisy label and $u_t$ is relatively small. In other words, the binary-reduced loss prevents the noisy labeled-class probability from increasing at the cost of substantially increasing the probabilities of other irrelevant classes. Moreover, by choosing different functions $g$, we can flexibly control how the loss responds to conditional probability values. The theorems further show that a broad class of univariate functions can be used for this purpose without compromising theoretical noise robustness.
	
	﻿

	\subsection{Instantiation of Robust Functions}
	
	The previous sections introduce two loss-construction schemes, Target Separation and Binary Reduction, both of which reduce robust loss design to the design of univariate functions. For both frameworks, the theoretical constraints on the underlying base functions can be reduced to conditions on their derivatives.
	Therefore, rather than designing the loss value directly, we design its derivative and recover the loss through integration. This provides a flexible way to construct admissible losses, especially when a simple closed-form antiderivative is unavailable.
	
	As examples, we consider two base derivative functions, 
	\begin{align}
		f_1(\alpha_1, \beta_1; x) &= \alpha_1 \cdot e^{\beta_1 (x - \gamma_1)^2}, \nonumber \\
		f_2(\alpha_2, \beta_2; x) &= \alpha_2  (x- \gamma_2)^2 +  \beta_2, \nonumber
	\end{align}
	where $\alpha_1, \beta_2  < 0, \alpha_2 \max \{ 1 - \gamma_2, \gamma_2 \}^2 + \beta_2 < 0$ and $ 0\leq \gamma_1, \gamma_2 \leq 1$. $f_1$ adopts a Gaussian-like exponential form, whose antiderivative can only be expressed using special functions (e.g., the error function family) rather than elementary closed-form expressions. In contrast, $f_2$ adopts a shifted quadratic form with a simple polynomial structure, whose antiderivative admits an elementary closed-form expression.
	
	Both $f_1$ and $f_2$ are strictly negative over $[0,1]$. Corollary~\ref{cor_1} indicates that the target-separatd loss is asymmetric under certain parameter settings. 
	The following derivation further characterizes this case.

	\begin{corollary}
		Under $f_1$ and $f_2$ as the base derivative functions, given weights $w_1, \ldots, w_K$ with $w_t > w_i, \forall i\neq t$ and $0 < w_t < 1$, the target-separatd loss $\mathcal{L}$ is asymmetric whenever any of the following conditions holds.
		
		(1) 
		\[
		\max_{i\neq t} \frac{w_t - w_i \xi }{(1-w_t) - (1-w_i)\xi  } \leq a \leq \frac{w_t}{1 - w_t}.
		\]
		
		(2) When $a > \frac{w_t}{1 - w_t}$, 
		\[
		\xi \leq \frac{(1-w_s)a - w_s}{(1-w_t)a - w_t}, \quad \forall s \neq t.
		\]
		
		(3)
		\[
		a = \frac{1}{K - 1}.
		\]
		
		For $f_1$ and $f_2$, the corresponding values of $\xi$ are $e^{|\beta_1|\max \{1-\gamma_1, \gamma_1\}^2}$ and $\max \{ \frac{\alpha_2}{\beta_2}\max \{1-\gamma_2,\gamma_2\}^2 + 1, (\frac{\alpha_2}{\beta_2}\max \{1-\gamma_2,\gamma_2\}^2 + 1)^{-1} \}$, respectively.
		
	\end{corollary}
	
	Consider a ten-class dataset with a symmetric noise rate of 0.8. 
	Setting $a = \frac{1}{30}$, we require that $\frac{0.8\xi/9 - 0.2}{8.2\xi/9 - 0.8} \leq \frac{1}{30}$, i.e., $1\leq \xi \leq \frac{234}{79}$.

	The binary-reduced loss induced by $f_1$ is symmetric when $\gamma_1 = 0.5$. Similarly, the binary-reduced loss induced by $f_2$ is symmetric when $\gamma_2 = 0.5$. For other choices of $\gamma_1$ and $\gamma_2$, the corresponding binary-reduced losses can be constructed according to the sufficient conditions given in Theorem~\ref{thm_5}.

	To further improve stability, we combine the target-separatd loss and binary-reduced loss into a unified formulation. It is defined as:
	\[
	\mathcal{L}_{TS+BR} = \lambda \cdot \mathcal{L}_{TS} + \mu \cdot \mathcal{L}_{BR}.
	\]
	We use $f_1$ and $f_2$ as the base derivative functions for both $\mathcal{L}_{TS}$ and $\mathcal{L}_{BR}$, and denote the resulting robust loss instances as BEF (Balanced Exponential Function) and BQF (Balanced Quadratic Function), respectively. Since the symmetric loss is asymmetric and the sum of asymmetric loss functions remains asymmetric~\cite{Zhou2021Asymmetric}, BEF and BQF are asymmetric and thus exhibits noise robustness.

	\begin{figure*}[htbp]
		\centering
		
		\subfloat[0.8 symmetric noise]{
			\includegraphics[width=0.24\linewidth]{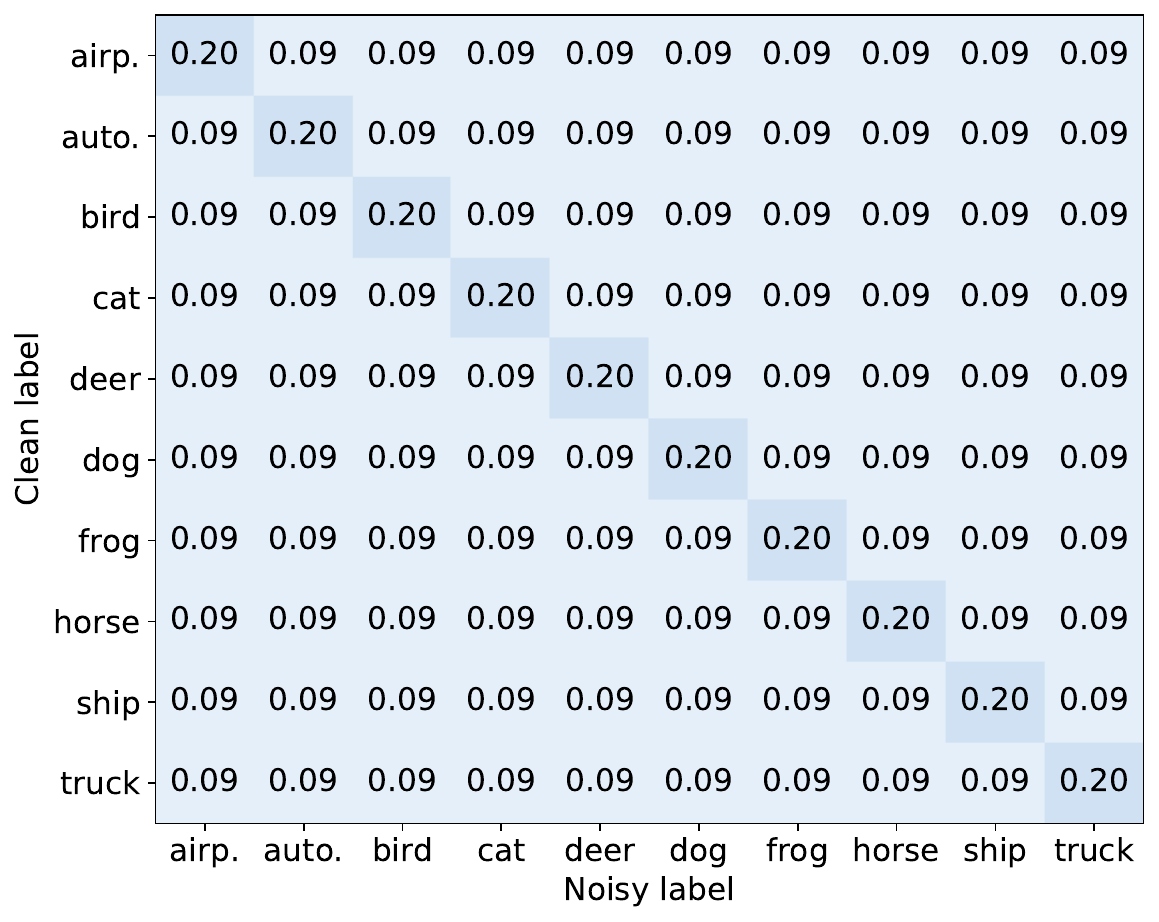}
		}
		\subfloat[0.4 asymmetric noise]{
			\includegraphics[width=0.24\linewidth]{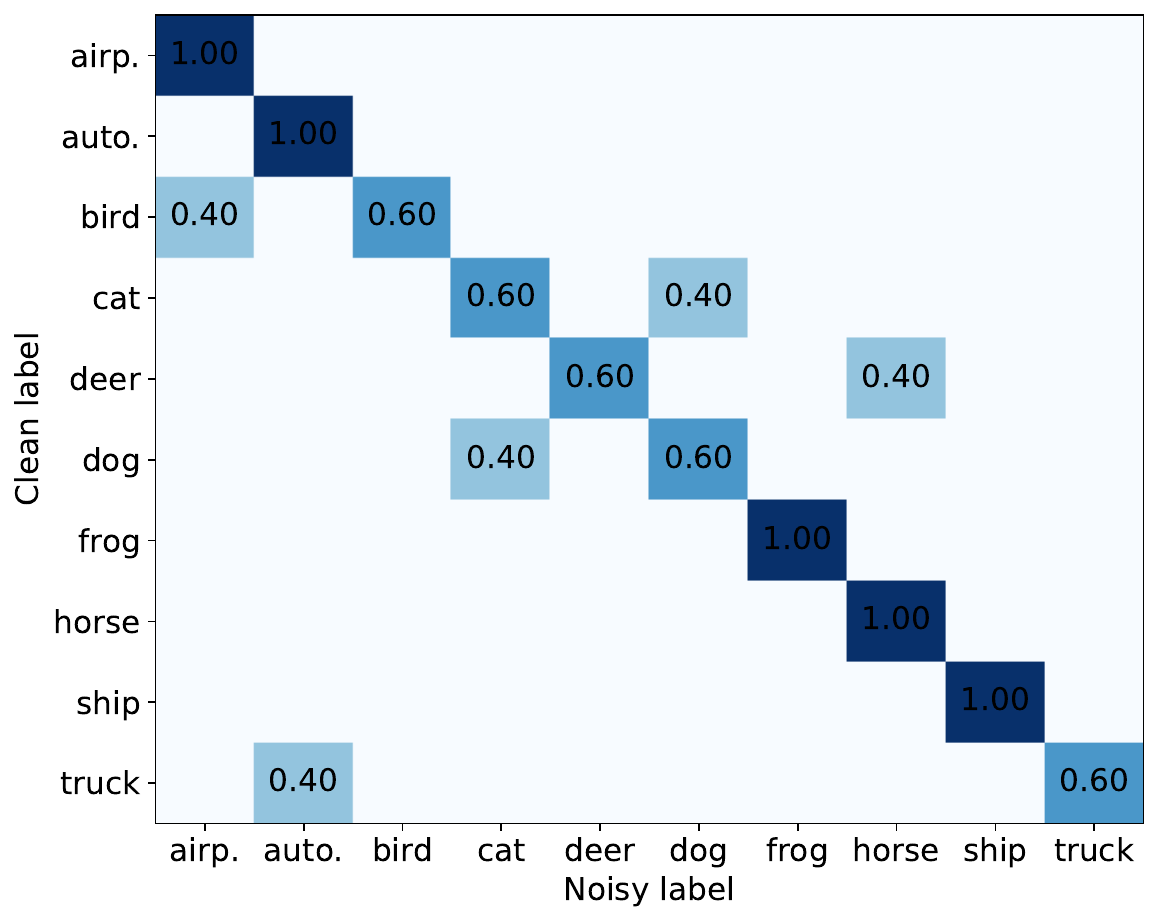}
		}
		\subfloat[0.6 dependent noise]{
			\includegraphics[width=0.24\linewidth]{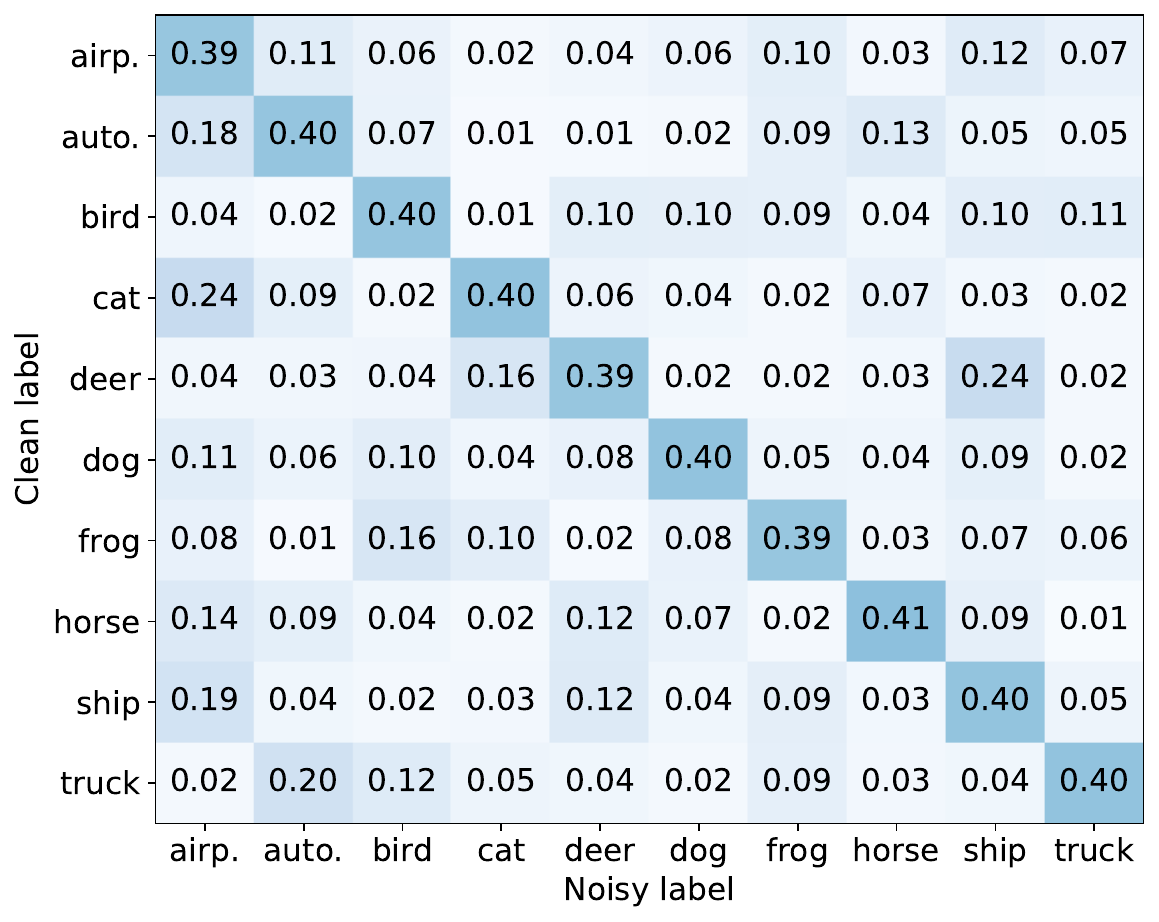}
		}
		\subfloat[Worst CIFAR-10N]{
			\includegraphics[width=0.24\linewidth]{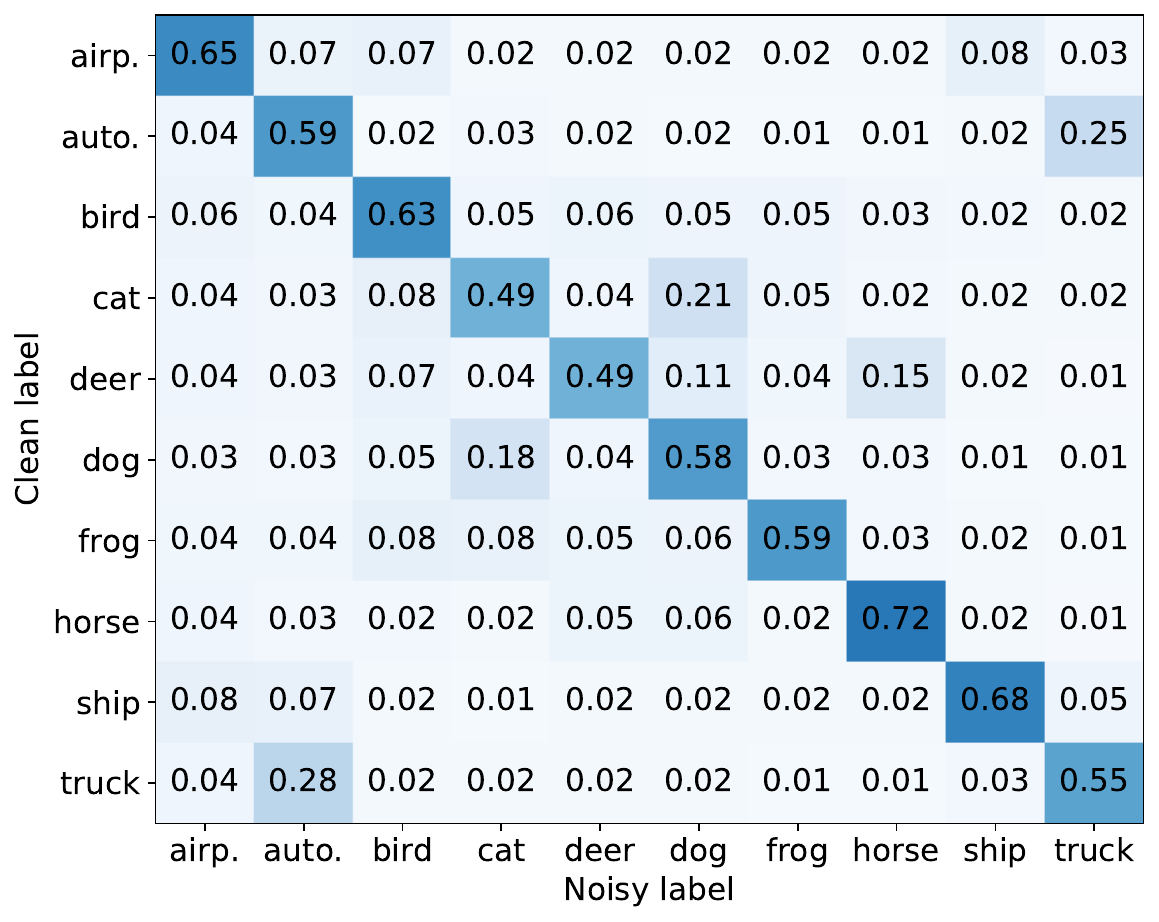}
		}

		\caption{
			Label transition matrices for representative noisy-label settings on CIFAR-10 and CIFAR-10N. The first three subfigures correspond to synthetic noise on CIFAR-10, including 0.8 symmetric noise, 0.4 asymmetric noise, and 0.6 instance-dependent noise. The last subfigure shows the Worst human-annotation setting from CIFAR-10N. Rows denote clean labels and columns denote noisy labels.
		}
		\label{fig_heatmap}
		﻿
	\end{figure*}

	\begin{table*}[t]
		\centering
		\setlength{\abovecaptionskip}{-0.03cm}
		\caption{Test accuracies (\%) of different methods on CIFAR-10 with clean, symmetric and asymmetric label noise. Results are reported as mean $\pm$ standard deviation over three random trials, with the top two results highlighted in bold.}
		\label{table_cifar10sa}
		\small
		\setlength{\tabcolsep}{4pt}
		\resizebox{\textwidth}{!}{
			\begin{tabular}{l|c|cccc|cccc}
				\toprule
				\multirow{2}{*}{\textbf{Loss}} 
				& \multirow{2}{*}{\textbf{Clean}} 
				& \multicolumn{4}{c|}{\textbf{Symmetric}} 
				& \multicolumn{4}{c}{\textbf{Asymmetric}} \\
				& & 0.2 & 0.4 & 0.6 & 0.8 
				& 0.1 & 0.2 & 0.3 & 0.4 \\
				\midrule
				
				CE & 90.44\std{0.06}
				& 74.81\std{0.30} & 58.07\std{0.19} & 38.80\std{0.30} & 19.19\std{0.16} 
				& 86.94\std{0.30} & 83.16\std{0.37} & 78.23\std{0.53} & 73.56\std{0.40} \\
				
				FL & 89.88\std{0.03}
				& 74.08\std{0.26} & 57.35\std{0.46} & 38.48\std{0.99} & 19.51\std{0.23} 
				& 86.50\std{0.17} & 83.10\std{0.26} & 78.42\std{0.44} & 73.74\std{0.36} \\
				
				GCE & 89.32\std{0.28}
				& 87.25\std{0.24} & 82.31\std{0.63} & 68.23\std{0.32} & 26.31\std{0.54} 
				& 88.28\std{0.26} & 85.57\std{0.19} & 79.58\std{0.48} & 72.77\std{0.27} \\
				
				SCE & 91.27\std{0.12}
				& 87.59\std{0.25} & 79.39\std{0.05} & 61.09\std{0.08} & 27.59\std{0.72} 
				& 89.53\std{0.39} & 86.03\std{0.20} & 80.40\std{0.23} & 73.99\std{0.37} \\
				
				NCE & 75.20\std{0.54}
				& 73.04\std{0.74} & 69.50\std{1.07} & 63.02\std{1.10} & 40.84\std{0.92} 
				& 74.07\std{0.57} & 72.22\std{0.35} & 69.44\std{0.32} & 65.44\std{0.61} \\

				CE+LC & 89.96\std{0.06}
				& 83.39\std{0.55} & 70.01\std{0.28} & 46.27\std{1.06} & 19.90\std{1.45} 
				& 87.35\std{0.14} & 83.49\std{0.21} & 78.85\std{0.50} & 73.33\std{0.09} \\
				
				$\text{CE}_\epsilon  \text{+MAE}$ & 91.08\std{0.17}
				& 89.17\std{0.27} & 85.97\std{0.30} & 79.84\std{0.37} & 56.69\std{1.41} 
				& 89.90\std{0.17} & 88.47\std{0.22} & 84.79\std{0.13} & 77.30\std{0.02} \\
				
				$\text{FL}_\epsilon  \text{+MAE}$ & 90.97\std{0.28}
				& 89.21\std{0.22} & 85.90\std{0.05} & 79.56\std{0.41} & 56.60\std{0.56} 
				& 90.01\std{0.19} & 88.29\std{0.53} & 84.84\std{0.24} & 77.03\std{0.15} \\
				
				NCE+RCE & 90.98\std{0.26}
				& 88.99\std{0.13} & 85.92\std{0.12} & 79.58\std{0.39} & 54.32\std{1.26} 
				& 90.19\std{0.10} & 88.30\std{0.10} & 85.20\std{0.52} & 78.03\std{0.51} \\
				
				NCE+AUL & 91.09\std{0.12}
				& 89.03\std{0.06} & 86.05\std{0.35} & 79.32\std{0.38} & 55.69\std{0.72} 
				& 90.10\std{0.33} & 88.39\std{0.18} & 84.63\std{0.50} & 77.23\std{0.33} \\
				
				NCE+AGCE & 90.91\std{0.05}
				& 89.18\std{0.06} & 86.11\std{0.39} & 79.73\std{0.45} & 45.27\std{5.04} 
				& 90.04\std{0.29} & 88.44\std{0.06} & 85.11\std{0.30} & 78.71\std{0.68} \\
				
				Anl-CE & 91.87\std{0.23}
				& 89.93\std{0.34} & 87.04\std{0.16} & 81.17\std{0.04} & 60.37\std{3.91} 
				& 90.78\std{0.16} & 89.13\std{0.08} & 85.75\std{0.28} & 77.94\std{0.21} \\
				
				Anl-FL & 91.59\std{0.15}
				& 89.91\std{0.11} & 87.35\std{0.42} & 81.44\std{0.18} & 61.72\std{3.30} 
				& 90.68\std{0.16} & 89.28\std{0.13} & 85.87\std{0.26} & 78.14\std{0.46} \\
				
				JAL-CE & 91.62\std{0.03}
				& 89.86\std{0.21} & 87.07\std{0.17} & 82.06\std{0.27} & 64.14\std{2.00} 
				& 90.61\std{0.10} & 89.27\std{0.06} & 86.52\std{0.16} & 79.73\std{0.44} \\
				
				JAL-FL & 91.76\std{0.33}
				& \textbf{90.33\std{0.18}} & 87.24\std{0.15} & 82.29\std{0.15} & 64.21\std{1.92}
				& 90.68\std{0.16} & 89.27\std{0.10} & 86.31\std{0.25} & 79.66\std{0.49} \\

				\midrule
				
				BEF
				& 91.63\std{0.14}
				& 90.03\std{0.17} & \textbf{87.45\std{0.09}} & \textbf{82.63\std{0.11}} & \textbf{64.54\std{2.09}}
				& \textbf{90.91\std{0.16}} & \textbf{89.77\std{0.19}} & \textbf{87.33\std{0.09}} & \textbf{80.73\std{0.24}}\\
				
				BQF
				& 91.56\std{0.32}
				& \textbf{90.23\std{0.19}} & \textbf{87.64\std{0.12}} & \textbf{82.73\std{0.21}} & \textbf{64.68\std{2.05}}
				& \textbf{90.80\std{0.21}} & \textbf{89.73\std{0.35}} & \textbf{87.20\std{0.10}} & \textbf{81.10\std{0.14}}\\

				\bottomrule
		\end{tabular}}
	\end{table*}

	\begin{table*}[t]
		\centering
		\setlength{\abovecaptionskip}{-0.03cm}
		\caption{Test accuracies (\%) of different methods on CIFAR-100 with clean, symmetric and asymmetric label noise. Results are reported as mean $\pm$ standard deviation over three random trials, with the top two results highlighted in bold.}
		\label{table_cifar100sa}
		\small
		\setlength{\tabcolsep}{4pt}
		\resizebox{\textwidth}{!}{
			\begin{tabular}{l|c|cccc|cccc}
				\toprule
				\multirow{2}{*}{\textbf{Loss}} 
				& \multirow{2}{*}{\textbf{Clean}} 
				& \multicolumn{4}{c|}{\textbf{Symmetric}} 
				& \multicolumn{4}{c}{\textbf{Asymmetric}} \\
				& & 0.2 & 0.4 & 0.6 & 0.8 
				& 0.1 & 0.2 & 0.3 & 0.4 \\
				\midrule

				CE & 71.44\std{0.22}
				& 56.96\std{0.41} & 40.01\std{1.53} & 23.00\std{1.39} & 7.77\std{0.15}
				& 63.63\std{1.60} & 58.36\std{0.88} & 50.38\std{0.39} & 41.81\std{0.31} \\
				
				FL & 70.94\std{0.74}
				& 56.61\std{0.72} & 40.90\std{1.31} & 22.98\std{0.91} & 7.96\std{0.57}
				& 63.95\std{0.65} & 58.04\std{0.93} & 50.56\std{0.60} & 41.56\std{0.23} \\
				
				GCE & 64.58\std{2.48}
				& 62.23\std{0.82} & 57.48\std{0.64} & 46.11\std{0.65} & 20.81\std{0.14}
				& 64.09\std{1.49} & 60.99\std{0.29} & 53.92\std{0.80} & 41.90\std{1.15} \\
				
				SCE & 70.58\std{0.34}
				& 56.40\std{0.48} & 38.75\std{0.57} & 22.48\std{0.83} & 7.90\std{0.14}
				& 64.50\std{0.48} & 57.58\std{0.76} & 50.03\std{1.24} & 41.64\std{0.74} \\
				
				NCE & 29.46\std{0.44}
				& 24.74\std{0.20} & 19.36\std{0.31} & 14.59\std{0.27} & 8.62\std{0.32}
				& 27.12\std{0.27} & 26.10\std{0.75} & 23.94\std{0.43} & 20.51\std{0.20} \\
				
				CE+LC & 70.88\std{0.28}
				& 57.08\std{0.82} & 37.90\std{0.49} & 17.70\std{0.08} & 6.78\std{0.38}
				& 63.29\std{0.28} & 55.92\std{0.40} & 47.81\std{0.15} & 39.39\std{0.26} \\
				
				$\text{CE}_\epsilon  \text{+MAE}$  & 69.95\std{0.85}
				& 65.39\std{0.70} & 59.30\std{0.35} & 49.22\std{1.75} & 26.32\std{0.81}
				& 66.97\std{0.38} & 62.86\std{0.42} & 59.25\std{2.02} & 51.02\std{0.51} \\
				
				$\text{FL}_\epsilon  \text{+MAE}$ & 70.25\std{0.46}
				& 65.69\std{1.32} & 58.46\std{0.54} & 48.66\std{1.53} & 26.61\std{1.59}
				& 67.13\std{0.54} & 64.38\std{0.09} & 58.14\std{0.41} & 50.53\std{0.02} \\
				
				NCE+RCE & 67.98\std{0.43}
				& 64.70\std{0.69} & 58.12\std{0.29} & 46.96\std{0.90} & 25.57\std{0.37}
				& 66.14\std{0.21} & 62.87\std{0.57} & 56.04\std{1.20} & 42.13\std{0.77} \\
				
				NCE+AUL & 69.65\std{0.33}
				& 65.07\std{0.36} & 56.74\std{0.30} & 38.89\std{0.20} & 13.12\std{0.43}
				& 66.12\std{0.16} & 57.89\std{0.70} & 48.32\std{0.45} & 38.81\std{0.16} \\
				
				NCE+AGCE & 69.00\std{0.18}
				& 65.19\std{0.37} & 59.46\std{0.47} & 47.73\std{0.86} & 25.24\std{1.68}
				& 67.05\std{0.50} & 64.08\std{0.78} & 56.88\std{0.44} & 44.38\std{0.62} \\
				
				Anl-CE & 70.38\std{0.50}
				& 67.25\std{0.93} & 61.81\std{0.66} & 51.97\std{0.41} & 28.31\std{0.46}
				& 68.66\std{0.39} & 65.91\std{0.29} & 60.06\std{0.76} & 45.53\std{0.61} \\
				
				Anl-FL & 70.39\std{0.28}
				& 67.32\std{0.38} & 61.50\std{0.42} & 51.48\std{0.39} & 27.36\std{1.60}
				& 68.40\std{0.30} & 66.16\std{0.13} & 60.21\std{0.91} & 45.76\std{0.46} \\

				JAL-CE & 70.65\std{0.10}
				& 68.16\std{0.29} & 64.35\std{0.70} & 56.29\std{0.89} & 22.89\std{1.45}
				& 69.17\std{0.29} & 67.60\std{0.59} & 64.70\std{0.60} & 56.72\std{0.93} \\
				
				JAL-FL & 70.42\std{0.11}
				& 67.90\std{0.22} & 64.48\std{0.16} & 56.30\std{0.21} & 21.43\std{2.06}
				& 69.15\std{0.50} & 68.13\std{0.51} & 65.42\std{0.37} & 55.67\std{0.77} \\
				
				\midrule
				
				BEF & 71.92\std{0.37}
				& \textbf{69.50\std{0.29}} & \textbf{64.62\std{0.26}} & \textbf{56.49\std{0.77}} & \textbf{32.81\std{0.66}}
				& \textbf{70.66\std{0.61}} & \textbf{69.38\std{0.18}} & \textbf{67.14\std{0.47}} & \textbf{59.19\std{0.47}} \\
				
				BQF & 71.46\std{0.30}
				& \textbf{69.15\std{0.23}} & \textbf{64.60\std{0.22}} & \textbf{57.02\std{0.75}} & \textbf{29.83\std{1.06}}
				& \textbf{70.17\std{0.15}} & \textbf{68.88\std{0.29}} & \textbf{66.67\std{0.44}} & \textbf{60.32\std{0.28}} \\

				\bottomrule
		\end{tabular}}
	\end{table*}

	\begin{table*}[t]
		\centering
		\setlength{\abovecaptionskip}{-0.03cm}
		\caption{Test accuracies (\%) of different methods on CIFAR-10/CIFAR-100 with instance-dependent noise (IDN) and human-annotated noise (CIFAR-10N/CIFAR-100N). Results are reported as mean $\pm$ standard deviation over three random trials, with the top two results in each column highlighted in bold.}
		\label{table_cifar_idn_human}
		\small
		\setlength{\tabcolsep}{4pt}
		\resizebox{\textwidth}{!}{
			\begin{tabular}{l|ccc|ccc|ccc|c}
				\toprule
				\multirow{2}{*}{\textbf{Loss}} 
				& \multicolumn{3}{c|}{\textbf{CIFAR-10 IDN}} 
				& \multicolumn{3}{c|}{\textbf{CIFAR-10N}} 
				& \multicolumn{3}{c|}{\textbf{CIFAR-100 IDN}} 
				& \textbf{CIFAR-100N} \\
				\cmidrule(lr){2-4} \cmidrule(lr){5-7} \cmidrule(lr){8-10} \cmidrule(lr){11-11}
				& 0.2 & 0.4 & 0.6 
				& Aggregate & Random 1 & Worst
				& 0.2 & 0.4 & 0.6
				& Noisy \\
				\midrule
				
				CE 
				& 74.79\std{0.26} & 57.92\std{0.49} & 37.24\std{0.20}
				& 85.14\std{0.13} & 78.83\std{0.39} & 61.82\std{0.24}
				& 55.86\std{1.05} & 41.94\std{0.69} & 24.71\std{0.87}
				& 49.68\std{0.17} \\
				
				FL 
				& 74.92\std{0.24} & 57.34\std{0.14} & 37.84\std{0.25}
				& 84.32\std{0.40} & 78.61\std{0.36} & 61.94\std{0.44}
				& 55.04\std{1.24} & 40.18\std{1.27} & 24.91\std{0.87}
				& 49.12\std{0.36} \\
				
				GCE 
				& 86.73\std{0.21} & 79.91\std{0.74} & 52.74\std{0.47}
				& 87.34\std{0.25} & 85.54\std{0.27} & 75.34\std{0.27}
				& 62.00\std{1.98} & 57.96\std{0.82} & 41.80\std{0.59}
				& 51.56\std{0.25} \\
				
				SCE 
				& 86.53\std{0.35} & 74.68\std{0.54} & 50.31\std{0.19}
				& 88.48\std{0.11} & 85.82\std{0.12} & 73.66\std{0.23}
				& 56.04\std{0.25} & 39.66\std{1.49} & 22.70\std{0.74}
				& 48.50\std{0.35} \\
				
				NCE 
				& 72.16\std{0.51} & 64.89\std{0.53} & 51.57\std{0.38}
				& 74.03\std{0.42} & 72.20\std{0.50} & 65.66\std{0.60}
				& 24.26\std{0.91} & 19.23\std{0.07} & 14.33\std{0.21}
				& 22.78\std{0.52} \\

				CE+LC 
				& 82.61\std{0.23} & 67.98\std{0.31} & 43.49\std{0.42}
				& 86.42\std{0.07} & 83.28\std{0.32} & 69.98\std{0.60}
				& 56.15\std{0.38} & 38.06\std{0.47} & 19.08\std{0.14}
				& 48.12\std{0.24} \\
				
				NCE+RCE 
				& 88.76\std{0.10} & 85.05\std{0.27} & 71.51\std{0.51}
				& 89.01\std{0.29} & 87.79\std{0.33} & 79.78\std{0.65}
				& 64.31\std{0.20} & 57.27\std{0.79} & 41.33\std{0.79}
				& 55.06\std{0.34} \\
				
				NCE+AUL 
				& 89.04\std{0.11} & 85.18\std{0.08} & 71.22\std{0.42}
				& 89.18\std{0.15} & 87.59\std{0.10} & 79.48\std{0.23}
				& 64.51\std{0.10} & 53.08\std{0.44} & 32.60\std{0.65}
				& 53.42\std{0.44} \\
				
				NCE+AGCE 
				& 88.92\std{0.10} & 85.17\std{0.59} & 73.10\std{0.46}
				& 88.95\std{0.15} & 88.07\std{0.12} & 80.01\std{0.21}
				& 65.51\std{0.68} & 58.57\std{0.14} & 42.92\std{0.46}
				& 55.68\std{0.20} \\

				Anl-CE 
				& 89.64\std{0.17} & 86.07\std{0.46} & 70.39\std{0.67}
				& 89.77\std{0.34} & 88.48\std{0.10} & 80.26\std{0.16}
				& 66.92\std{0.31} & 61.26\std{0.97} & 47.56\std{0.05}
				& 56.78\std{0.48} \\
				
				Anl-FL 
				& 89.62\std{0.21} & 86.42\std{0.22} & 70.63\std{0.41}
				& 89.88\std{0.23} & 88.31\std{0.22} & 79.99\std{0.19}
				& 66.66\std{0.29} & 61.25\std{0.06} & 47.28\std{0.93}
				& 56.99\std{0.47} \\

				JAL-CE 
				& 89.77\std{0.42} & 86.52\std{0.11} & 75.46\std{0.33}
				& \textbf{90.04\std{0.19}} & 88.85\std{0.19} & 81.16\std{0.20}
				& 68.10\std{0.32} & 63.64\std{0.16} & 51.71\std{0.42}
				& 59.33\std{0.17} \\
				
				JAL-FL
				& \textbf{89.85\std{0.11}} & 86.66\std{0.21} & 75.15\std{0.22}
				& 89.76\std{0.20} & 88.86\std{0.17} & 81.15\std{0.27}
				& 68.05\std{0.05} & 63.67\std{0.24} & 51.86\std{0.99}
				& 59.44\std{0.22} \\
				
				\midrule
				
				BEF
				& \textbf{90.00\std{0.20}} & \textbf{86.76\std{0.14}} & \textbf{76.11\std{0.26}}
				& \textbf{90.10\std{0.08}} & \textbf{89.00\std{0.13}} & \textbf{81.95\std{0.46}}
				& \textbf{68.81\std{0.24}} & \textbf{64.56\std{0.14}} & \textbf{54.70\std{0.60}}
				& \textbf{59.98\std{0.60}} \\
				
				BQF
				& 89.69\std{0.17} & \textbf{86.72\std{0.09}} & \textbf{75.80\std{0.39}}
				& 89.82\std{0.33} & \textbf{88.93\std{0.19}} & \textbf{81.76\std{0.15}}
				& \textbf{68.62\std{0.33}} & \textbf{64.05\std{0.30}} & \textbf{53.51\std{0.27}}
				& \textbf{59.81\std{0.19}} \\
				
				\bottomrule
			\end{tabular}
		}
	\end{table*}

	\begin{table*}[t]
		\centering
		\setlength{\abovecaptionskip}{-0.03cm}
		\caption{Test accuracies (\%) of BEF/BQF under shared parameter configurations and their task-specific tuned variants (BEF+/BQF+) on CIFAR-10 and CIFAR-100 with severe symmetric, asymmetric, instance-dependent, and human-annotated noise.}
		\label{table_cifar_best}
		\small
		\setlength{\tabcolsep}{4pt}
		\resizebox{0.98\textwidth}{!}{
			\begin{tabular}{l|cccc|cccc}
				\toprule
				\multirow{2}{*}{\textbf{Loss}} 
				& \multicolumn{4}{c|}{\textbf{CIFAR-10 }} 
				& \multicolumn{4}{c}{\textbf{CIFAR-100}}  \\
				\cmidrule(lr){2-5} \cmidrule(lr){6-9} 
				& Sym. 0.8 & Asy. 0.4 & IDN 0.6 
				& Worst & Sym. 0.8 & Asy. 0.4 & IDN 0.6 
				& Noisy\\
				\midrule
				
				BEF
				& \textbf{64.54\std{2.09}} & 80.73\std{0.24} & 76.11\std{0.26} & 81.95\std{0.46} 
				& \textbf{32.81\std{0.66}}  & 59.19\std{0.47} & 54.70\std{0.60} & 59.98\std{0.60}
				\\
				BEF+
				& \textbf{64.54\std{2.09}} & \textbf{84.02\std{0.15}} & \textbf{77.40\std{0.53}} & \textbf{82.35\std{0.08}}
				& \textbf{32.81\std{0.66}} & \textbf{61.55\std{0.09}} & \textbf{54.85\std{0.25}} & \textbf{61.02\std{0.32}}
				\\
				\midrule
				
				BQF 
				& \textbf{64.68\std{2.05}} & 81.10\std{0.14} & 75.80\std{0.39} & 81.76\std{0.15} 
				& 29.83\std{1.06} & 60.32\std{0.28} & 53.51\std{0.27} & 59.81\std{0.19}
				\\
				BQF+ 
				& \textbf{64.68\std{2.05}} & \textbf{84.17\std{0.41}} & \textbf{77.86\std{0.57}} & \textbf{82.25\std{0.23}}
				& \textbf{31.75\std{1.07}} & \textbf{61.76\std{0.48}} & \textbf{55.00\std{0.54}} & \textbf{60.67\std{0.14}}
				\\
				\bottomrule
			\end{tabular}
		}
	\end{table*}

	\begin{table*}[htbp]
		\centering
		\setlength{\abovecaptionskip}{-0.03cm}
		\caption{Test accuracies (\%) of different methods on WebVision, ILSVRC12, and Clothing1M, including Top-1 and Top-5 results. The best two results are highlighted in bold.}
		\label{table_realworld}
		\small
		\setlength{\tabcolsep}{3pt}
		\resizebox{\textwidth}{!}{
			\begin{tabular}{llcccccccccccc}
				\toprule
				& & CE & FL & GCE & $\text{CE}_\epsilon+\text{MAE}$ & $\text{FL}_\epsilon+\text{MAE}$ & NCE+RCE & NCE+AGCE & JAL-CE & JAL-FL & BEF & BQF \\
				\midrule
				\multirow{2}{*}{\textbf{WebVision}} 
				& Top-1 & 65.48 & 66.56 & 69.92 & 70.32 & 70.24 & 69.20 & 67.68 & 67.76 & 68.16 & \textbf{72.16} & \textbf{71.08} \\
				& Top-5 & 85.40 & 86.00 & 88.28 & 87.32 & 88.36 & 86.44 & 85.60 & 86.08 & 86.09 & \textbf{89.36} & \textbf{88.56} \\
				\midrule
				\multirow{2}{*}{\textbf{ILSVRC12}} 
				& Top-1 & 61.36 & 61.04 & 65.96 & 65.32 & 66.84 & 65.24 & 64.12 & 65.68 & 63.84 & \textbf{68.64} & \textbf{68.24} \\
				& Top-5 & 85.08 & 85.36 & 87.92 & 87.64 & 87.20 & 85.48 & 85.00 & 86.12 & 84.64 & \textbf{89.20} & \textbf{88.64} \\
				\midrule
				\multirow{2}{*}{\textbf{Clothing1M}} 
				& Top-1 & 67.95 & 68.00 & 69.07 & 69.68 & 69.87 & 69.58 & 68.31 & 69.51 & 69.78 & \textbf{69.97} & \textbf{69.89} \\
				& Top-5 & 98.25 & 98.46 & \textbf{98.54} & 98.35 & 98.33 & 98.03 & 97.78 & 98.47 & 97.87 & 97.99 & \textbf{98.61} \\
				
				\bottomrule
		\end{tabular}}
	\end{table*}

	\begin{figure}[htbp]
		\centering
		
		\subfloat[CE]{
			\includegraphics[width=0.48\linewidth]{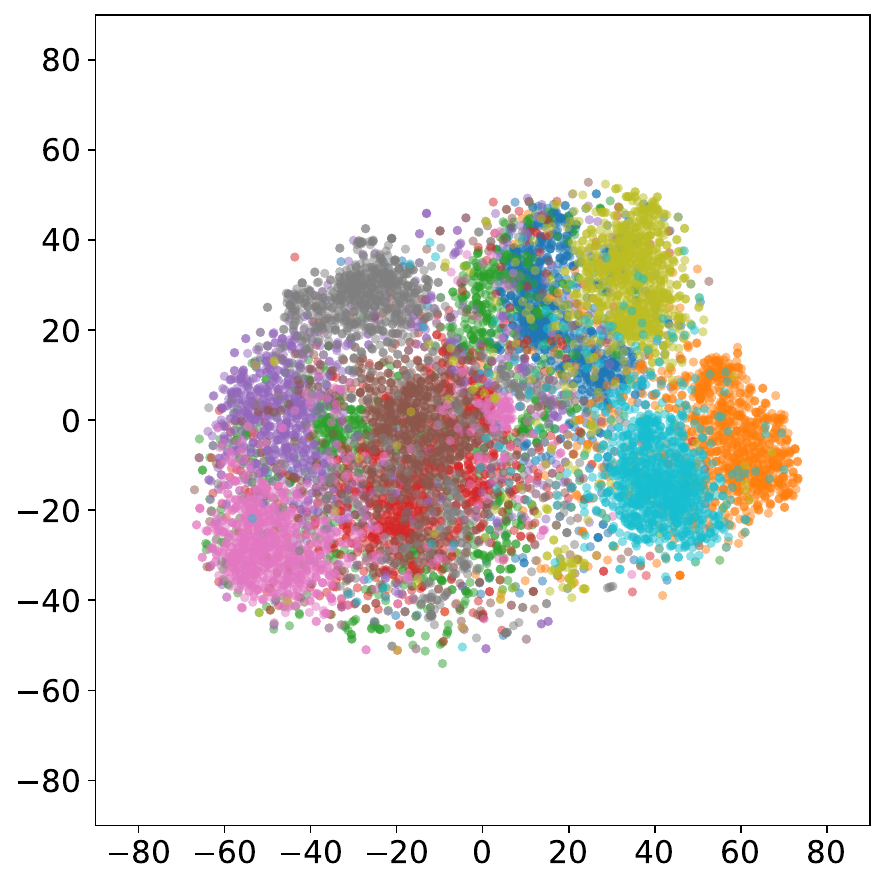}
			\label{fig_tsneCE}
		}
		\subfloat[FL]{
			\includegraphics[width=0.48\linewidth]{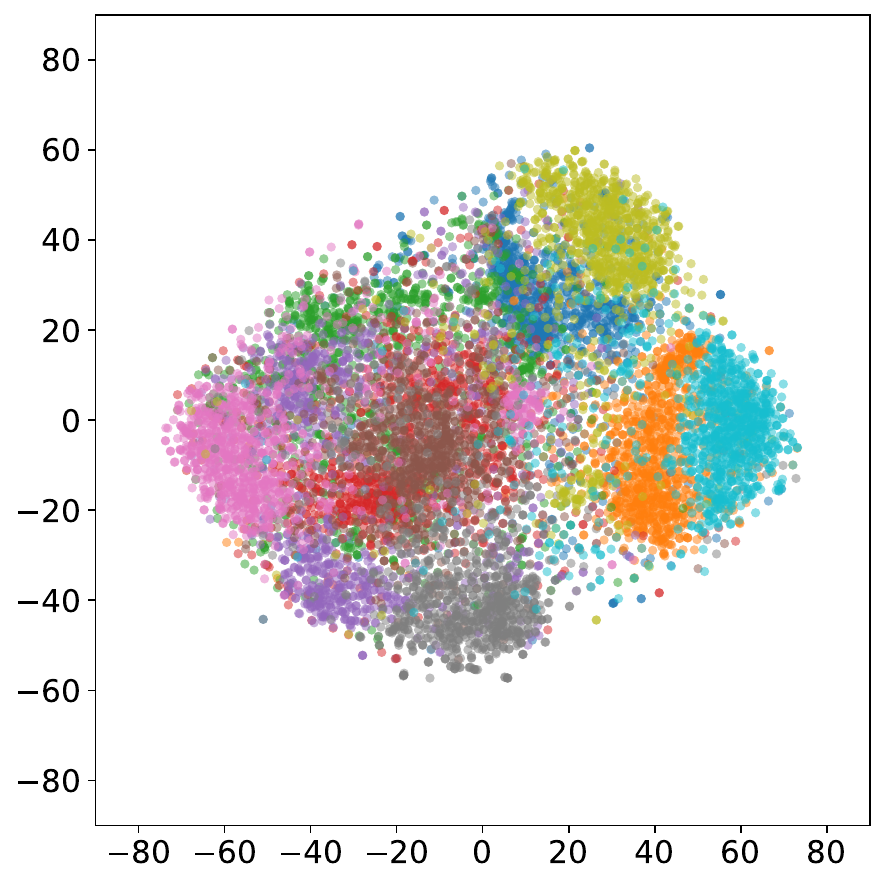}
			\label{fig_tsneFL}
		}
		
		\subfloat[BEF]{
			\includegraphics[width=0.48\linewidth]{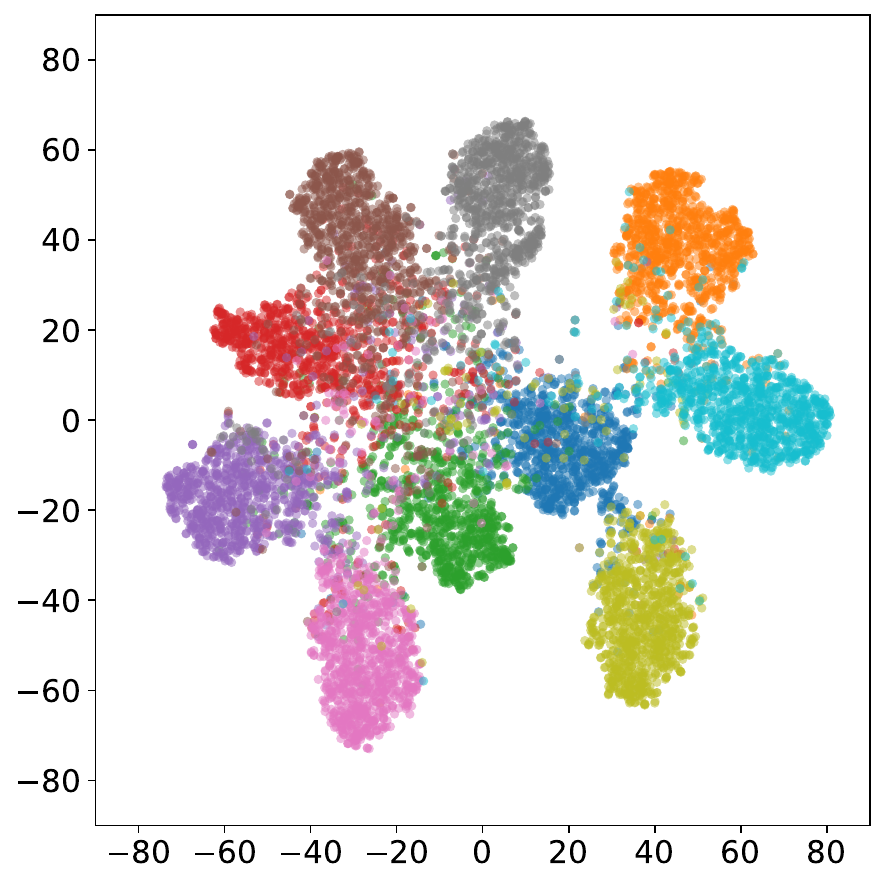}
			\label{fig_tsneBEF}
		}
		\subfloat[BQF]{
			\includegraphics[width=0.48\linewidth]{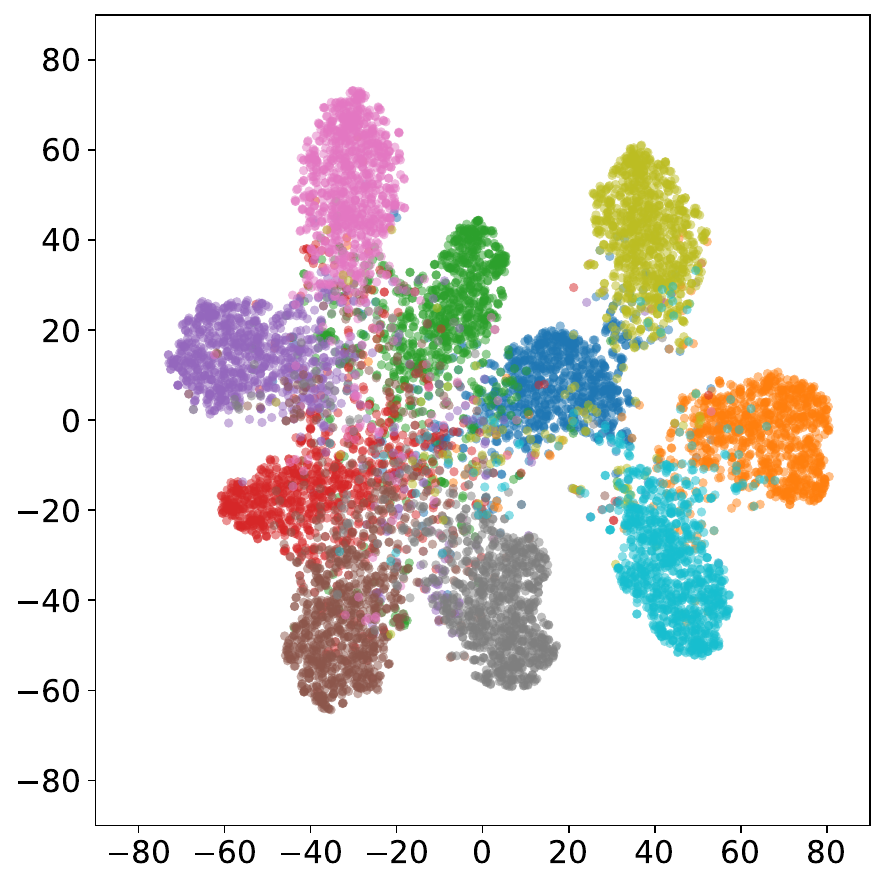}
			\label{fig_tsneBQF}
		}
		
		\caption{2D t-SNE \cite{Laurens2008TSNE} visualization of learned representations on the CIFAR-10 test set under 0.4 instance-dependent noise. BEF and BQF yield more discriminative feature distributions with improved class separation and clearer boundaries.}
		\label{fig_TSNE}
	\end{figure}

	\begin{figure*}[htbp]
		\centering
		
		\subfloat[0.8 symmetric noise]{
			\includegraphics[width=0.24\linewidth]{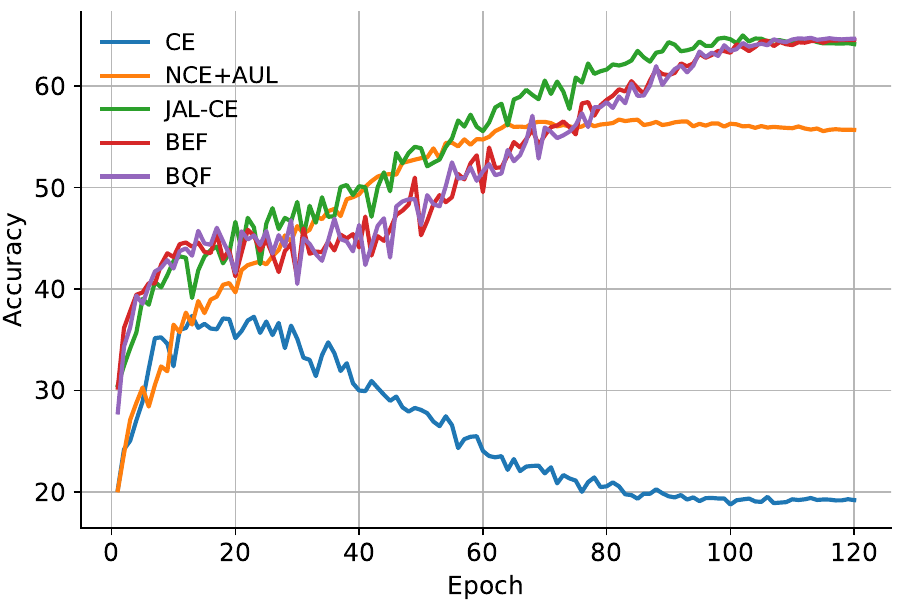}
			\label{fig_cifar10selected_acca}
		}
		\subfloat[0.4 asymmetric noise]{
			\includegraphics[width=0.24\linewidth]{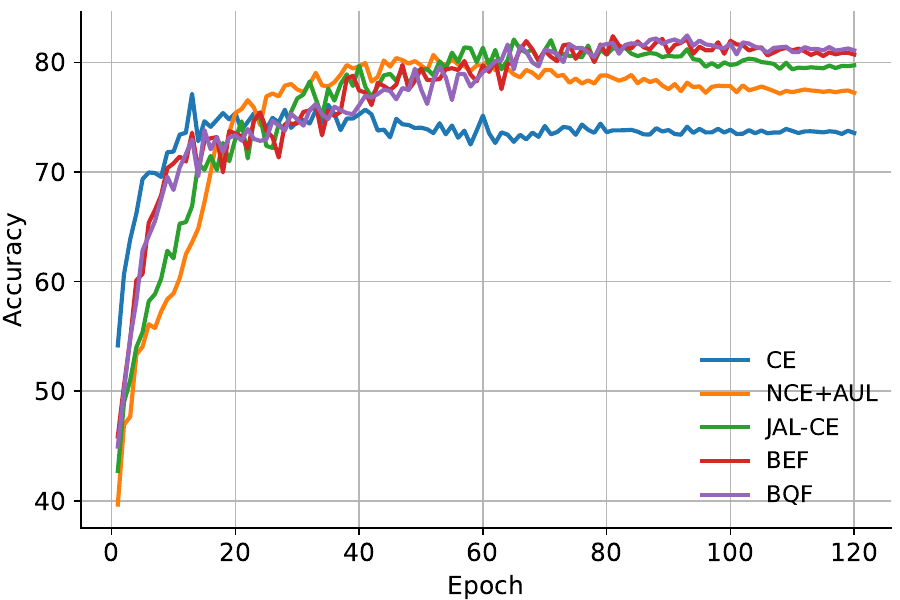}
			\label{fig_cifar10selected_accb}
		}
		\subfloat[0.6 dependent noise]{
			\includegraphics[width=0.24\linewidth]{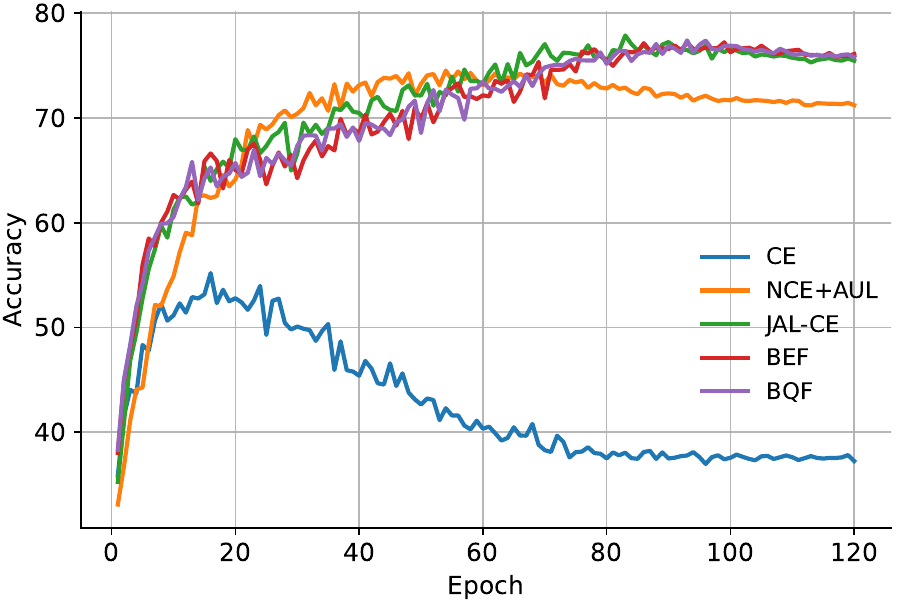}
			\label{fig_cifar10selected_accc}
		}
		\subfloat[Worst CIFAR-10N]{
			\includegraphics[width=0.24\linewidth]{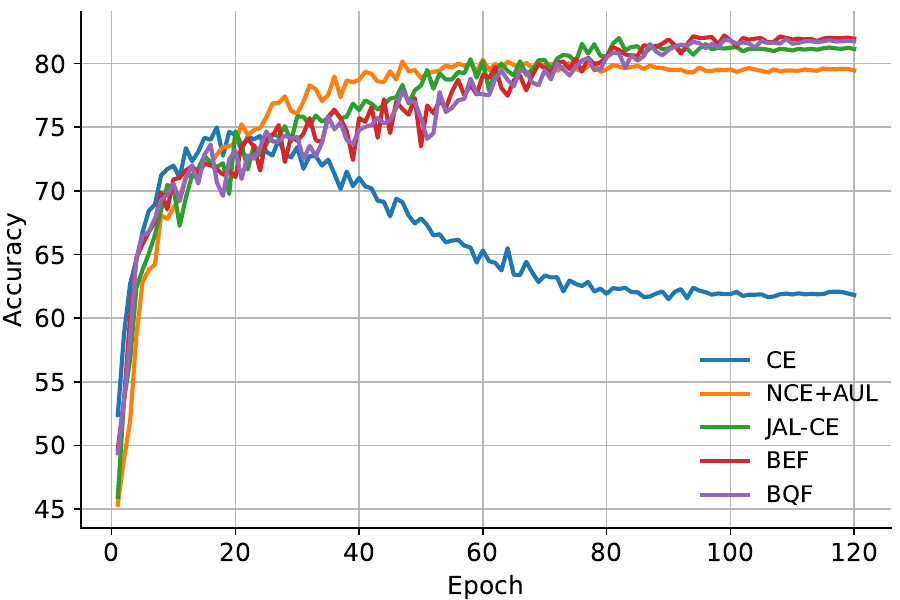}
			\label{fig_cifar10selected_accd}
		}

		\caption{Test accuracy curves of different methods under representative noisy settings on CIFAR-10.}
		\label{fig_cifar10selected_acc}
	\end{figure*}

	\begin{figure*}[htbp]
		\centering
		
		\subfloat[0.8 symmetric noise]{
			\includegraphics[width=0.24\linewidth]{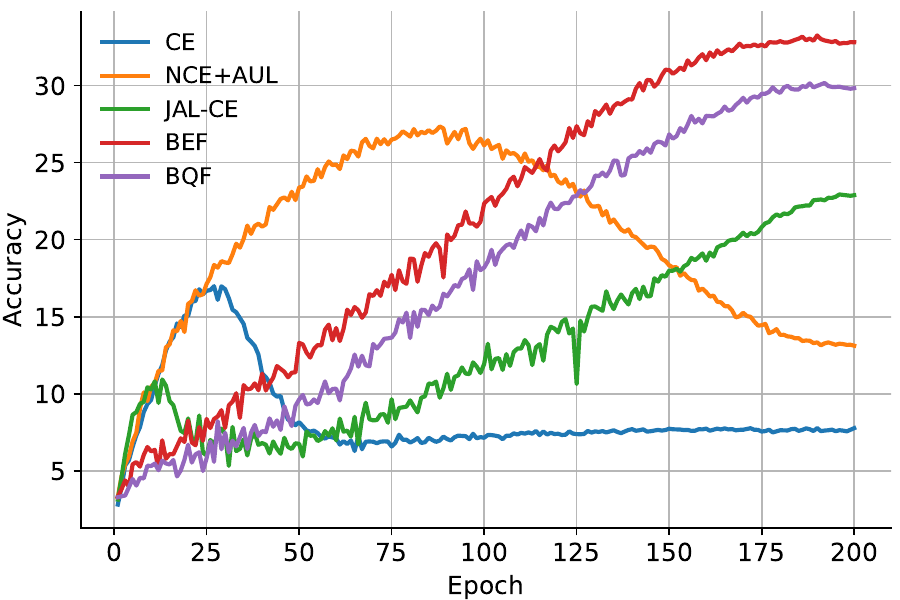}
			\label{fig_cifar100selected_acca}
		}
		\subfloat[0.4 asymmetric noise]{
			\includegraphics[width=0.24\linewidth]{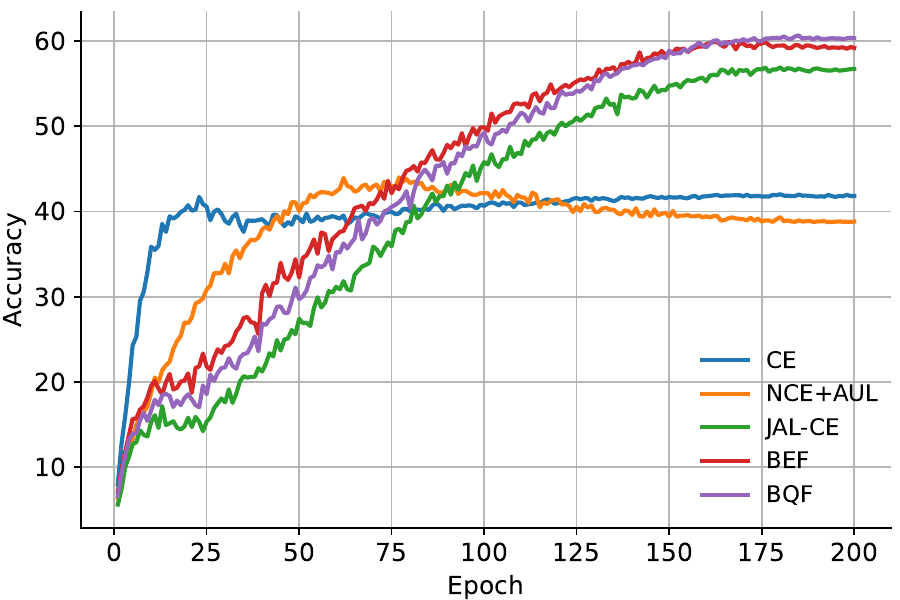}
			\label{fig_cifar100selected_accb}
		}
		\subfloat[0.6 dependent noise]{
			\includegraphics[width=0.24\linewidth]{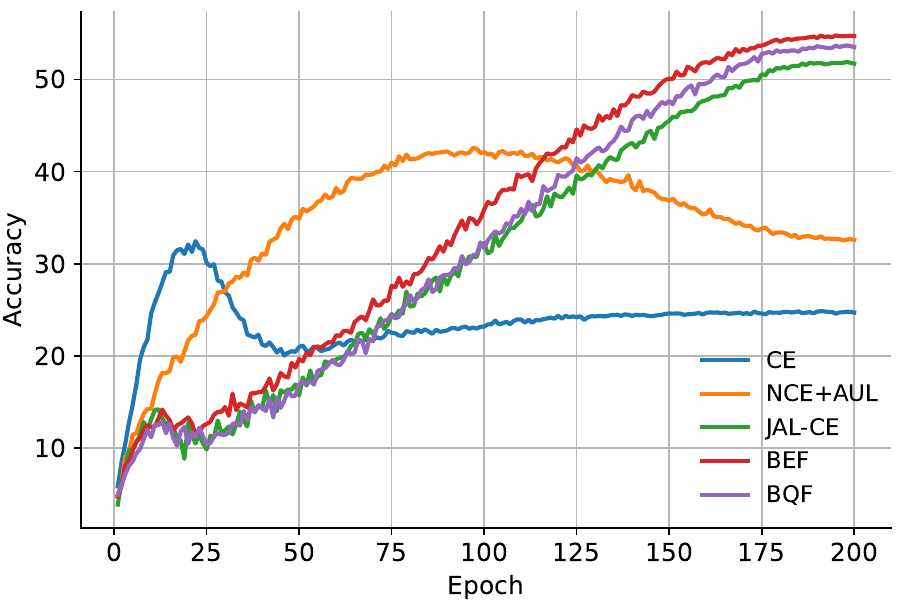}
			\label{fig_cifar100selected_accc}
		}
		\subfloat[Noisy CIFAR-100N]{
			\includegraphics[width=0.24\linewidth]{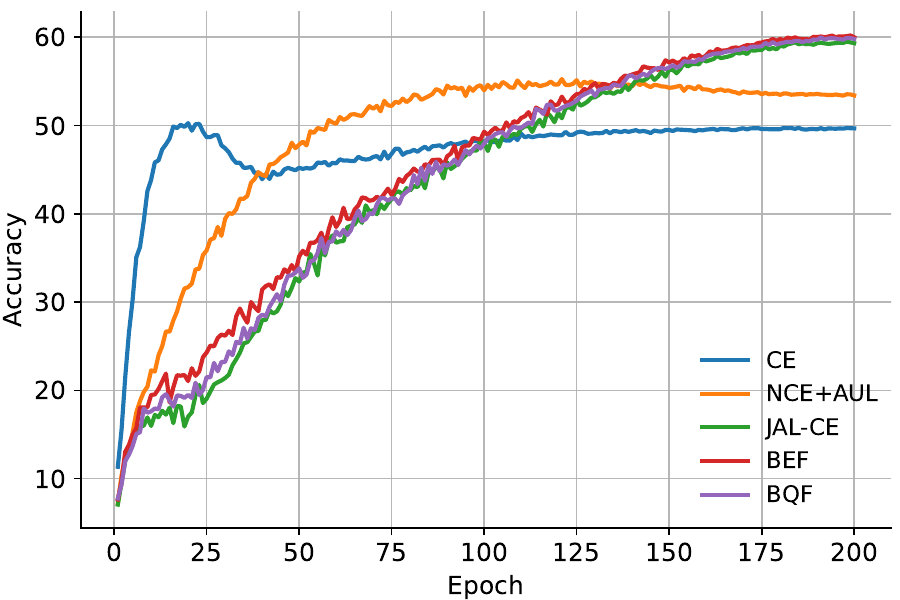}
			\label{fig_cifar100selected_accd}
		}

		\caption{Test accuracy curves of different methods under representative noisy settings on CIFAR-100.}
		\label{fig_cifar100selected_acc}
	\end{figure*}

	\begin{figure*}[htbp]
		\centering
		
		\subfloat[BEF on CIFAR-10N]{
			\includegraphics[width=0.23\linewidth]{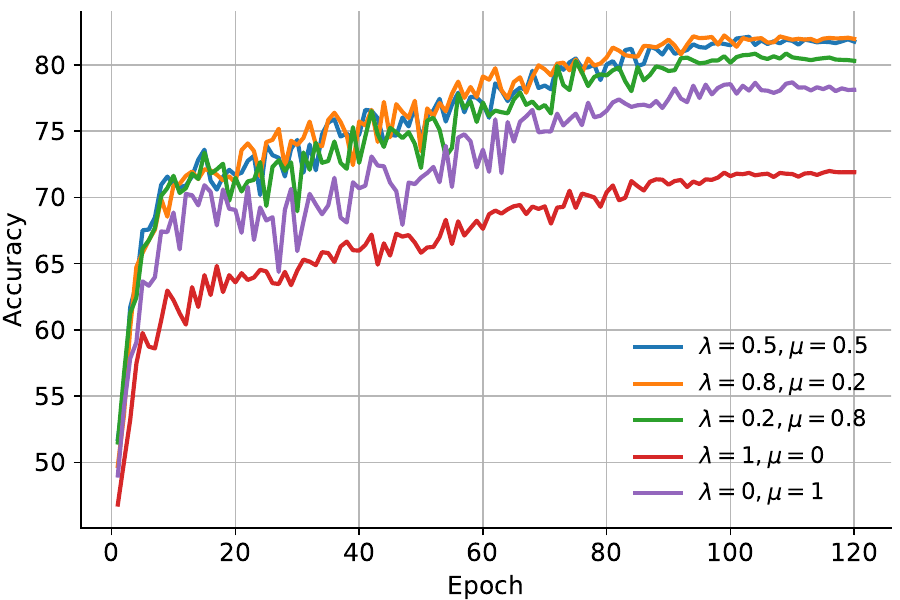}
			\label{fig_weighta}
		}
		\subfloat[BQF on CIFAR-10N]{
			\includegraphics[width=0.23\linewidth]{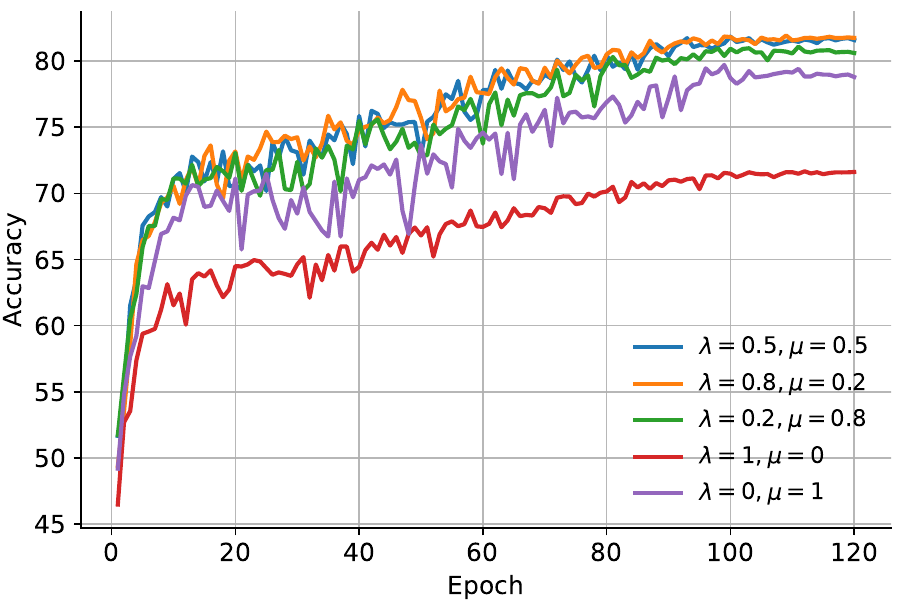}
			\label{fig_weightb}
		}
		\subfloat[BEF on CIFAR-100N]{
			\includegraphics[width=0.23\linewidth]{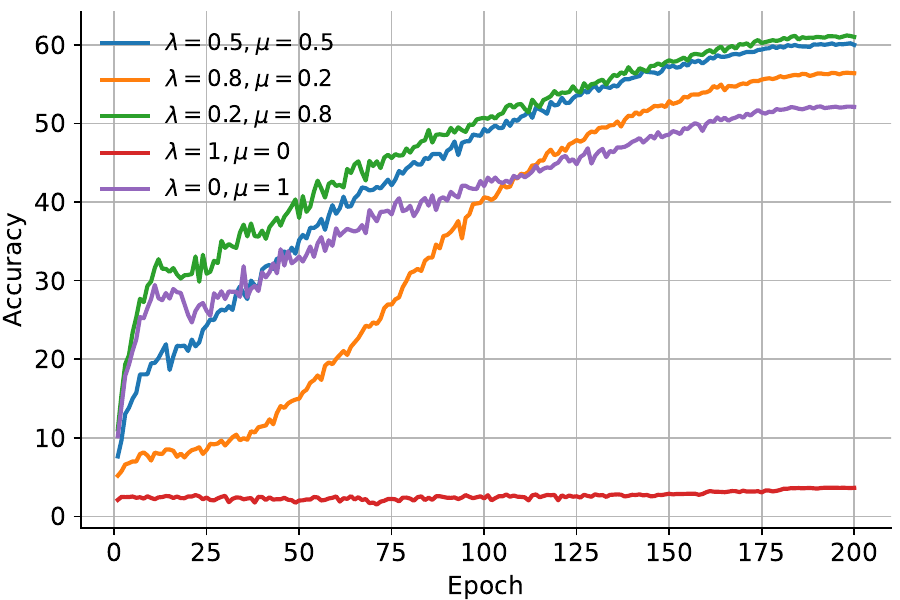}
			\label{fig_weightc}
		}
		\subfloat[BQF on CIFAR-100N]{
			\includegraphics[width=0.23\linewidth]{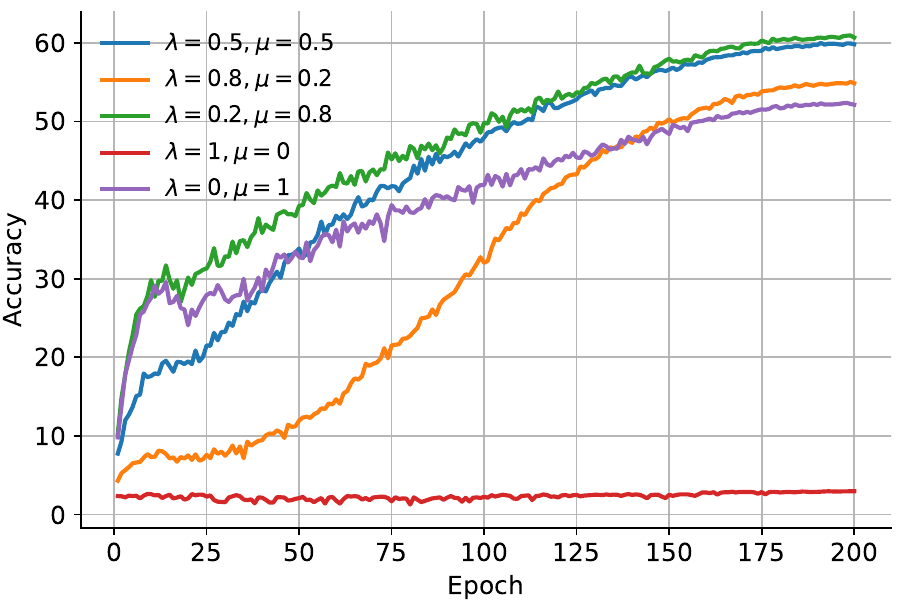}
			\label{fig_weightd}
		}			
		\caption{Test accuracies of BEF and BQF with different weighting coefficients under ``Worst'' noise types on CIFAR-10N and ``Noisy'' noise types on CIFAR-100N.}
		\label{fig_weight}
	\end{figure*}
	
	\section{Experiments}
	\subsection{Experimental Setup}
	We consider image classification tasks and evaluate the performance on seven popular datasets. CIFAR-10 and CIFAR-100~\cite{Krizhevsky2009cifar} are used for 10- and 100-class natural image classification, respectively, while their noisy counterparts, CIFAR-10N and CIFAR-100N~\cite{wei2022cifarn}, introduce real-world human-annotated label noise. For large-scale evaluation, we further consider WebVision~\cite{Li2017WebVision}, a web-crawled noisy dataset, with ILSVRC12 also serving as the clean counterpart. In addition, Clothing1M~\cite{Tong2015Clothing1M} is included as a large-scale real-world noisy-label benchmark containing one million images from 14 categories.
	
	For CIFAR-10 and CIFAR-100, we conduct experiments under both synthetic and human-annotated noisy-label settings. Specifically, we consider three types of synthetic noise, including symmetric noise, asymmetric noise, and instance-dependent noise. Symmetric noise uniformly corrupts labels to other classes, while asymmetric noise flips labels according to a predefined class-transition matrix. Instance-dependent noise corrupts labels based on both the input instance and the ground-truth class through a class-conditional label generator. In addition, CIFAR-10N and CIFAR-100N are used to evaluate real-world human annotation noise. Fig.~\ref{fig_heatmap} visualizes representative label transition matrices on CIFAR-10 and CIFAR-10N, including symmetric, asymmetric, instance-dependent, and human noise. These matrices provide an intuitive comparison of the noise patterns considered in our experiments.
	
	Following prior work~\cite{Wang2025JAL}, we adopt dataset-specific architectures. 
	For CIFAR-10 and CIFAR-100, we use an 8-layer CNN~\cite{Yann1989CNN} and ResNet-34\cite{He2016ResNet}, respectively. 
	For WebVision, we follow the commonly used mini setting~\cite{Lu2018Mentornet} by selecting the first 50 classes from the Google image subset, training a ResNet-50, and evaluating on the corresponding 50 classes of both the WebVision and ILSVRC12 validation sets. 
	For Clothing1M, we initialize a ResNet-50 with ImageNet pretraining, train on the noisy training set, and evaluate on the test set.
	
	We compare the proposed method with a broad set of representative baselines, including Cross Entropy (CE), Focal Loss (FL)~\cite{Lin2017FL}, Generalized Cross Entropy (GCE)~\cite{Zhang2018GCE}, Symmetric Cross Entropy (SCE)\cite{Wang2019SCE}, LogitClip (CE+LC)~\cite{Wei2023CELC}, and $\epsilon$-softmax-enhanced losses ($\text{CE}_\epsilon+\text{MAE}$ and $\text{FL}_\epsilon+\text{MAE}$)~\cite{Wang2024epsilon}. We also include normalization-based losses, such as Active Passive Loss (APL, including NCE and NCE+RCE)~\cite{Ma2020NCERCE}, Asymmetric Loss Functions (ALFs, including NCE+AUL and NCE+AGCE)~\cite{Zhou2021Asymmetric,Zhou2023Asymmetric}, Active Negative Loss (ANL, including ANL-CE and ANL-FL)~\cite{Ye2023ANL}, and Joint Asymmetric Loss (JAL, including JAL-CE and JAL-FL)~\cite{Wang2025JAL}. More details are provided in the Appendix. The implementation will be made publicly available.

	\subsection{Evaluation on Benchmark Datasets}

	Tables~\ref{table_cifar10sa}--\ref{table_cifar_idn_human} summarize the results on CIFAR-10, CIFAR-100, CIFAR-10N, and CIFAR-100N under both synthetic and real-world noisy settings. Overall, the proposed methods achieve consistently competitive performance across different datasets and noise types, validating the effectiveness of the proposed robust loss construction framework.
	
	The BEF and BQF perform particularly well under challenging noisy-label settings, including asymmetric noise, instance-dependent noise, and human-annotated noise, where stronger robustness is generally required. Their advantages are especially evident on CIFAR-100 and real-world noisy benchmarks, suggesting that the proposed framework remains effective even when the noise structure becomes more complex. 
	
	Specifically, under symmetric and asymmetric label noise on CIFAR-10, BEF and BQF rank among the top-performing methods in most settings.
	In particular, under 0.8 symmetric noise, BQF obtains 64.68\% accuracy and BEF obtains 64.54\%, both slightly outperforming JAL-FL.
	Under asymmetric noise, the proposed methods show clearer advantages, with BQF achieving 81.10\% accuracy under 0.4 asymmetric noise.
	
	On CIFAR-100, the proposed methods also perform competitively under both symmetric and asymmetric noise.
	For symmetric noise, BEF and BQF obtain the top two results in all noise ratios, with BQF achieving the highest accuracy under 0.6 noise and BEF performing best under 0.2, 0.4, and 0.8 noise.
	For asymmetric noise, BEF and BQF consistently achieve the best two results across all corruption levels, and BQF reaches 60.32\% accuracy under 0.4 asymmetric noise.
	
	For instance-dependent noise and human-annotated noise, BEF and BQF further demonstrate strong robustness.
	On CIFAR-10, BEF achieves the best results under all three IDN settings and also obtains the highest accuracies on the Aggregate, Random 1 and Worst subsets of CIFAR-10N.
	On CIFAR-100, BEF ranks first under all three IDN settings and CIFAR-100N.
	These results suggest that the proposed losses remain effective not only under synthetic noise, but also under realistic noisy-label scenarios.
	
	Fig.~\ref{fig_TSNE} further presents the 2D t-SNE visualization of learned representations under 0.4 instance-dependent noise on CIFAR-10. 
	Compared with CE and FL, BEF and BQF produce more discriminative feature distributions with clearer class separation and decision boundaries, which is consistent with their improved robustness under noisy supervision.  
	Figs.~\ref{fig_cifar10selected_acc} and~\ref{fig_cifar100selected_acc} illustrate the test accuracy of different methods under various noise types. On CIFAR-10, CE shows clear overfitting under strong symmetric and instance-dependent noise, whereas robust methods remain more stable. BEF and BQF achieve competitive or superior performance across all settings. On CIFAR-100, their advantages are even more pronounced under severe noise, demonstrating both stronger final accuracy and more stable optimization.
	These results demonstrate that the proposed losses generalize well beyond synthetic noise assumptions and remain effective under realistic noisy-label scenarios.
	
	Overall, the experimental results demonstrate the effectiveness of combining target separation and binary reduction objectives. In particular, although BEF and BQF do not rely on the normalization-based formulations adopted by methods such as NCE+RCE~\cite{Ma2020NCERCE}, NCE+AUL~\cite{Zhou2021Asymmetric}, NCE+AGCE~\cite{Zhou2021Asymmetric}, ANL-CE~\cite{Ye2023ANL}, ANL-FL~\cite{Ye2023ANL}, JAL-CE~\cite{Wang2025JAL}, and JAL-FL~\cite{Wang2025JAL}, they still achieve competitive and often superior performance under challenging noisy settings. This suggests that robust loss design can also benefit from the alternative perspective of combining target separation and binary reduction, beyond the normalization-based paradigm.

	\subsection{Effect of Task-Specific Parameter Tuning}
	﻿
	We further examine whether task-specific tuning can improve the proposed losses beyond their shared parameter configurations. In the main experiments, BEF and BQF use largely shared settings across different noise types and noise rates, which provides a simple and general-purpose choice. Table~\ref{table_cifar_best} compares these shared configurations with their tuned variants, denoted as BEF+ and BQF+, under challenging noise settings.
	
	The results show that task-specific tuning improves BEF and BQF in most cases, especially under structured noise such as asymmetric and instance-dependent noise. On CIFAR-10, BEF+ improves BEF by 3.29 and 1.29 percentage points under 0.4 asymmetric noise and 0.6 instance-dependent noise, respectively, while BQF+ yields corresponding gains of 3.07 and 2.06 points. On CIFAR-100, BEF+ improves BEF from 59.19\% to 61.55\% under 0.4 asymmetric noise, and BQF+ improves BQF from 53.51\% to 55.00\% under 0.6 instance-dependent noise. Similar gains are also observed on the human-annotated noise settings.
	
	These results indicate that the shared configurations already provide strong and stable performance, while task-specific tuning can further adapt the proposed losses to more challenging noise patterns. Therefore, BEF and BQF can be used either as general-purpose robust losses with shared settings or as flexible loss families that can be further optimized for specific datasets and noise conditions.
	﻿

	\subsection{Evaluation on Real-World Datasets}

	Table~\ref{table_realworld} reports the classification accuracies on WebVision, ILSVRC12, and Clothing1M, which contain large-scale real-world noisy labels. It presents the Top-1 results together with supplementary Top-5 results. Overall, BEF and BQF achieve consistently competitive performance across all datasets, demonstrating the scalability of the proposed framework beyond small-scale synthetic benchmarks. In particular, the strong results on both WebVision and Clothing1M indicate that the proposed losses remain effective even when the noise distribution is more complex and less controlled. These findings suggest that the combination of target separation and binary reduction generalizes well to practical noisy-label learning scenarios.
	
	On the WebVision benchmark, BEF achieves the highest accuracy of 72.16\%, outperforming all compared baselines by a clear margin. 
	A similar trend is observed on the corresponding ILSVRC12 validation set, where BEF reaches 68.64\% accuracy, indicating improved transferability and robustness under cross-dataset evaluation. 
	BQF also achieves competitive performance on both datasets, further validating the effectiveness of the proposed framework.
	
	On Clothing1M, which contains about one million real-world noisy images, BEF and BQF achieve 69.97\% and 69.89\% accuracy, respectively, showing competitive performance compared with existing robust loss functions. 
	These results suggest that the proposed methods remain effective not only under synthetic noise assumptions, but also in large-scale real-world noisy-label settings.

	\subsection{Ablation Study on Target Separation and Binary Reduction}
	
	Similar to many recent robust loss functions, our method combines multiple components to improve robustness against noisy labels. In this experiment, we study the effect of the weighting coefficients assigned to target separation and binary reduction. Fig.~\ref{fig_weight} presents the test accuracy curves of BEF and BQF under different coefficient settings on CIFAR-10N and CIFAR-100N.
	
	As shown in Fig.~\ref{fig_weight}, different choices of $(\lambda,\mu)$ lead to noticeably different optimization trajectories and final accuracies. On CIFAR-10N, coefficient settings that jointly incorporate both target separation and binary reduction, such as $(0.8,0.2)$ and $(0.5,0.5)$, consistently achieve strong performance for both BEF and BQF, while relying solely on target separation or binary reduction is generally less effective. On CIFAR-100N, this trend becomes even more pronounced: the setting $(1,0)$ performs poorly throughout training, whereas combined settings such as $(0.2,0.8)$ and $(0.5,0.5)$ yield substantially better convergence and final accuracy.
	
	These results suggest that target separation and binary reduction provide complementary benefits during optimization. In most cases, appropriately balancing the two objectives leads to more stable training and improved robustness under noisy supervision.

	\section{Conclusion}

	In this work, we presented a general framework for robust loss construction based on univariate base functions. The proposed framework induces multiclass losses through two complementary formulations: Target Separation and Binary Reduction. Target Separation constructs an inter-class independent loss by decoupling target and non-target contributions, while Binary Reduction introduces inter-class dependencies through normalized pairwise comparisons. For both formulations, we analyzed symmetry and asymmetry properties and derived corresponding sufficient conditions, providing theoretical guidance for robust loss design. Based on the framework, we further instantiated two robust losses and evaluated them on synthetic and real-world noisy-label benchmarks. The experimental results demonstrate the effectiveness of the proposed losses under diverse noise settings. Overall, this work provides a flexible and theoretically grounded perspective for designing robust losses in noisy-label learning.

	\bibliographystyle{IEEEtran}
	\bibliography{references}

@article{wang2026variation,
  title={Variation-Bounded Loss for Noise-Tolerant Learning},
  author={Wang, Jialiang and Zhou, Xiong and Liu, Xianming and Hu, Gangfeng and Zhai, Deming and Jiang, Junjun and Li, Haoliang},
  journal={Proceedings of the AAAI Conference on Artificial Intelligence},
  volume={40},
  number={31},
  pages={26251--26259},
  year={2026},
  publisher={Association for the Advancement of Artificial Intelligence},
}

@inproceedings{Zhang2018GCE,
 author = {Zhang, Zhilu and Sabuncu, Mert},
 booktitle = {Advances in Neural Information Processing Systems},
 pages = {},
 title = {Generalized Cross Entropy Loss for Training Deep Neural Networks with Noisy Labels},
 volume = {31},
 year = {2018}
}

@INPROCEEDINGS{Wang2019SCE,
  author={Wang, Yisen and Ma, Xingjun and Chen, Zaiyi and Luo, Yuan and Yi, Jinfeng and Bailey, James},
  booktitle={2019 IEEE/CVF International Conference on Computer Vision (ICCV)}, 
  title={Symmetric Cross Entropy for Robust Learning With Noisy Labels}, 
  year={2019},
  volume={},
  number={},
  pages={322-330}
}

@InProceedings{Ma2020NCERCE,
  title = 	 {Normalized Loss Functions for Deep Learning with Noisy Labels},
  author =       {Ma, Xingjun and Huang, Hanxun and Wang, Yisen and Romano, Simone and Erfani, Sarah and Bailey, James},
  booktitle = 	 {Proceedings of the 37th International Conference on Machine Learning},
  pages = 	 {6543--6553},
  year = 	 {2020},
  volume = 	 {119},
  series = 	 {Proceedings of Machine Learning Research},
  month = 	 {13--18 Jul}
}

@ARTICLE{Zhang2024LTAPL,
  author={Zhang, Shuo and Li, Jian-Qing and Fujita, Hamido and Li, Yu-Wen and Wang, Deng-Bao and Zhu, Ting-Ting and Zhang, Min-Ling and Liu, Cheng-Yu},
  journal={IEEE Transactions on Pattern Analysis and Machine Intelligence}, 
  title={Student Loss: Towards the Probability Assumption in Inaccurate Supervision}, 
  year={2024},
  volume={46},
  number={6},
  pages={4460-4475}
}

@inproceedings{Ye2023ANL,
 author = {Ye, Xichen and Li, Xiaoqiang and dai, songmin and Liu, Tong and Sun, Yan and Tong, Weiqin},
 booktitle = {Advances in Neural Information Processing Systems},
 pages = {6917--6940},
 title = {Active Negative Loss Functions for Learning with Noisy Labels},
 volume = {36},
 year = {2023}
}

@INPROCEEDINGS{Lin2017FL,
  author={Lin, Tsung-Yi and Goyal, Priya and Girshick, Ross and He, Kaiming and Dollár, Piotr},
  booktitle={2017 IEEE International Conference on Computer Vision (ICCV)}, 
  title={Focal Loss for Dense Object Detection}, 
  year={2017},
  volume={},
  number={},
  pages={2999-3007}
}

@InProceedings{Zhou2021Asymmetric,
  title = 	 {Asymmetric Loss Functions for Learning with Noisy Labels},
  author =       {Zhou, Xiong and Liu, Xianming and Jiang, Junjun and Gao, Xin and Ji, Xiangyang},
  booktitle = 	 {Proceedings of the 38th International Conference on Machine Learning},
  pages = 	 {12846--12856},
  year = 	 {2021},
  volume = 	 {139},
  series = 	 {Proceedings of Machine Learning Research},
  month = 	 {18--24 Jul}
}

@ARTICLE{Zhou2023Asymmetric,
  author={Zhou, Xiong and Liu, Xianming and Zhai, Deming and Jiang, Junjun and Ji, Xiangyang},
  journal={IEEE Transactions on Pattern Analysis and Machine Intelligence}, 
  title={Asymmetric Loss Functions for Noise-Tolerant Learning: Theory and Applications}, 
  year={2023},
  volume={45},
  number={7},
  pages={8094-8109}
}

@InProceedings{Wei2023CELC,
  title = 	 {Mitigating Memorization of Noisy Labels by Clipping the Model Prediction},
  author =       {Wei, Hongxin and Zhuang, Huiping and Xie, Renchunzi and Feng, Lei and Niu, Gang and An, Bo and Li, Yixuan},
  booktitle = 	 {Proceedings of the 40th International Conference on Machine Learning},
  pages = 	 {36868--36886},
  year = 	 {2023},
  volume = 	 {202},
  series = 	 {Proceedings of Machine Learning Research},
  month = 	 {23--29 Jul}
}

@article{Aritra2017Robust, 
title={Robust Loss Functions under Label Noise for Deep Neural Networks}, 
volume={31}, 
number={1}, 
journal={Proceedings of the AAAI Conference on Artificial Intelligence}, 
author={Ghosh, Aritra and Kumar, Himanshu and Sastry, P. S.}, 
year={2017}, 
month={Feb.} }

@ARTICLE{Naresh2013Tolerance,
  author={Manwani, Naresh and Sastry, P. S.},
  journal={IEEE Transactions on Cybernetics}, 
  title={Noise Tolerance Under Risk Minimization}, 
  year={2013},
  volume={43},
  number={3},
  pages={1146-1151}
  }

@inproceedings{van2015Learning,
 author = {van Rooyen, Brendan and Menon, Aditya and Williamson, Robert C},
 booktitle = {Advances in Neural Information Processing Systems},
 pages = {},
 title = {Learning with Symmetric Label Noise: The Importance of Being Unhinged},
 volume = {28},
 year = {2015}
}

@inproceedings{Lei2020Taylor,
  title     = {Can Cross Entropy Loss Be Robust to Label Noise?},
  author    = {Feng, Lei and Shu, Senlin and Lin, Zhuoyi and Lv, Fengmao and Li, Li and An, Bo},
  booktitle = {Proceedings of the Twenty-Ninth International Joint Conference on
               Artificial Intelligence, {IJCAI-20}},
  publisher = {International Joint Conferences on Artificial Intelligence Organization},
  editor    = {Christian Bessiere},
  pages     = {2206--2212},
  year      = {2020},
  month     = {7},
  note      = {Main track}
}

@inproceedings{Erik2021JS,
 author = {Englesson, Erik and Azizpour, Hossein},
 booktitle = {Advances in Neural Information Processing Systems},
 pages = {30284--30297},
 title = {Generalized Jensen-Shannon Divergence Loss for Learning with Noisy Labels},
 volume = {34},
 year = {2021}
}

@INPROCEEDINGS{Zhou2021SR,
  author={Zhou, Xiong and Liu, Xianming and Wang, Chenyang and Zhai, Deming and Jiang, Junjun and Ji, Xiangyang},
  booktitle={2021 IEEE/CVF International Conference on Computer Vision (ICCV)}, 
  title={Learning with Noisy Labels via Sparse Regularization}, 
  year={2021},
  volume={},
  number={},
  pages={72-81}
}

@inproceedings{Wang2024epsilon,
 author = {Wang, Jialiang and Zhou, Xiong and Zhai, Deming and Jiang, Junjun and Ji, Xiangyang and Liu, Xianming},
 booktitle = {Advances in Neural Information Processing Systems},
 pages = {32012--32038},
 title = {{$\epsilon$}-Softmax: Approximating One-Hot Vectors for Mitigating Label Noise},
 volume = {37},
 year = {2024}
}

@inproceedings{Wang2025JAL,
  title     = {{Joint Asymmetric Loss for Learning with Noisy Labels}},
  author    = {Wang, Jialiang and Liu, Xianming and Zhou, Xiong and Hu, Gangfeng and Zhai, Deming and Jiang, Junjun and Ji, Xiangyang},
  booktitle = {International Conference on Computer Vision},
  year      = {2025},
  pages     = {1947-1956}
}

@InProceedings{Wei2022To,
  title = 	 {To Smooth or Not? {W}hen Label Smoothing Meets Noisy Labels},
  author =       {Wei, Jiaheng and Liu, Hangyu and Liu, Tongliang and Niu, Gang and Sugiyama, Masashi and Liu, Yang},
  booktitle = 	 {Proceedings of the 39th International Conference on Machine Learning},
  pages = 	 {23589--23614},
  year = 	 {2022},
  volume = 	 {162},
  series = 	 {Proceedings of Machine Learning Research},
  month = 	 {17--23 Jul}
}

@inproceedings{Menon2020PHuber,
title={Can gradient clipping mitigate label noise?},
author={Aditya Krishna Menon and Ankit Singh Rawat and Sashank J. Reddi and Sanjiv Kumar},
booktitle={International Conference on Learning Representations},
year={2020},
}

@INPROCEEDINGS{Kim2019NLNL,
  author={Kim, Youngdong and Yim, Junho and Yun, Juseung and Kim, Junmo},
  booktitle={2019 IEEE/CVF International Conference on Computer Vision (ICCV)}, 
  title={{NLNL}: Negative Learning for Noisy Labels}, 
  year={2019},
  volume={},
  number={},
  pages={101-110}
}

@INPROCEEDINGS{Kim2021JNPL,
  author={Kim, Youngdong and Yun, Juseung and Shon, Hyounguk and Kim, Junmo},
  booktitle={2021 IEEE/CVF Conference on Computer Vision and Pattern Recognition (CVPR)}, 
  title={Joint Negative and Positive Learning for Noisy Labels}, 
  year={2021},
  volume={},
  number={},
  pages={9437-9446}
}

@inproceedings{Xia2020Part,
 author = {Xia, Xiaobo and Liu, Tongliang and Han, Bo and Wang, Nannan and Gong, Mingming and Liu, Haifeng and Niu, Gang and Tao, Dacheng and Sugiyama, Masashi},
 booktitle = {Advances in Neural Information Processing Systems},
 pages = {7597--7610},
 title = {Part-dependent Label Noise: Towards Instance-dependent Label Noise},
 volume = {33},
 year = {2020}
}

@InProceedings{Charoenphakdee2019On,
  title = 	 {On Symmetric Losses for Learning from Corrupted Labels},
  author =       {Charoenphakdee, Nontawat and Lee, Jongyeong and Sugiyama, Masashi},
  booktitle = 	 {Proceedings of the 36th International Conference on Machine Learning},
  pages = 	 {961--970},
  year = 	 {2019},
  volume = 	 {97},
  series = 	 {Proceedings of Machine Learning Research},
  month = 	 {09--15 Jun}
}

@inproceedings{Krizhevsky2009cifar,
  title={Learning Multiple Layers of Features from Tiny Images},
  author={Alex Krizhevsky},
  year={2009}
}

@inproceedings{wei2022cifarn,
title={Learning with Noisy Labels Revisited: A Study Using Real-World Human Annotations},
author={Jiaheng Wei and Zhaowei Zhu and Hao Cheng and Tongliang Liu and Gang Niu and Yang Liu},
booktitle={International Conference on Learning Representations},
year={2022}
}

@article{Li2017WebVision,
      title={WebVision Database: Visual Learning and Understanding from Web Data}, 
      author={Wen Li and Limin Wang and Wei Li and Eirikur Agustsson and Luc Van Gool},
      journal      = {arXiv preprint arXiv:1708.02862},
      year         = {2017},
      eprinttype    = {arXiv},
      eprint       = {1708.02862}
}

@INPROCEEDINGS{Tong2015Clothing1M,
  author={Tong Xiao and Tian Xia and Yi Yang and Chang Huang and Xiaogang Wang},
  booktitle={2015 IEEE Conference on Computer Vision and Pattern Recognition (CVPR)}, 
  title={Learning from massive noisy labeled data for image classification}, 
  year={2015},
  volume={},
  number={},
  pages={2691-2699}
}

@ARTICLE{Yann1989CNN,
  author={LeCun, Yann and Boser, Bernhard and Denker, John S Denker and Henderson, Donnie and Richard E Howard and Hubbard, Wayne and Lawrence D Jackel},
  journal={Neural Computation}, 
  title={Backpropagation Applied to Handwritten Zip Code Recognition}, 
  year={1989},
  volume={1},
  number={4},
  pages={541-551}
  }

@INPROCEEDINGS{He2016ResNet,
  author={He, Kaiming and Zhang, Xiangyu and Ren, Shaoqing and Sun, Jian},
  booktitle={2016 IEEE Conference on Computer Vision and Pattern Recognition (CVPR)}, 
  title={Deep Residual Learning for Image Recognition}, 
  year={2016},
  volume={},
  number={},
  pages={770-778}
}

@article{Bartlett2006Risk,
author = {Peter L Bartlett and Michael I Jordan and Jon D McAuliffe},
title = {Convexity, Classification, and Risk Bounds},
journal = {Journal of the American Statistical Association},
volume = {101},
number = {473},
pages = {138--156},
year = {2006}
}

@article{Yann2015Deep,
  author  = {LeCun, Yann and Bengio, Yoshua and Hinton, Geoffrey},
  title   = {Deep learning},
  journal = {Nature},
  year    = {2015},
  volume  = {521},
  number  = {7553},
  pages   = {436--444}
}

@article{Bo2021Survey,
      title={A Survey of Label-noise Representation Learning: Past, Present and Future}, 
      author={Bo Han and Quanming Yao and Tongliang Liu and Gang Niu and Ivor W. Tsang and James T. Kwok and Masashi Sugiyama},
    journal      = {arXiv preprint arXiv:2011.04406},
      year={2021},
     eprinttype    = {arXiv},
      eprint={2011.04406}
}

@InProceedings{Devansh2017Closer,
  title = 	 {A Closer Look at Memorization in Deep Networks},
  author =       {Devansh Arpit and Stanis{\l}aw Jastrz{\k{e}}bski and Nicolas Ballas and David Krueger and Emmanuel Bengio and Maxinder S. Kanwal and Tegan Maharaj and Asja Fischer and Aaron Courville and Yoshua Bengio and Simon Lacoste-Julien},
  booktitle = 	 {Proceedings of the 34th International Conference on Machine Learning},
  pages = 	 {233--242},
  year = 	 {2017},
  volume = 	 {70},
  series = 	 {Proceedings of Machine Learning Research},
  month = 	 {06--11 Aug}
}

@InProceedings{Collin2024Weak,
  title = 	 {Weak-to-Strong Generalization: Eliciting Strong Capabilities With Weak Supervision},
  author =       {Burns, Collin and Izmailov, Pavel and Kirchner, Jan Hendrik and Baker, Bowen and Gao, Leo and Aschenbrenner, Leopold and Chen, Yining and Ecoffet, Adrien and Joglekar, Manas and Leike, Jan and Sutskever, Ilya and Wu, Jeffrey},
  booktitle = 	 {Proceedings of the 41st International Conference on Machine Learning},
  pages = 	 {4971--5012},
  year = 	 {2024},
  volume = 	 {235},
  series = 	 {Proceedings of Machine Learning Research},
  month = 	 {21--27 Jul}
}

@article{Laurens2008TSNE,
  author  = {Laurens van der Maaten and Geoffrey Hinton},
  title   = {Visualizing Data using t-SNE},
  journal = {Journal of Machine Learning Research},
  year    = {2008},
  volume  = {9},
  number  = {86},
  pages   = {2579--2605}
}

@InProceedings{Lu2018Mentornet,
  title = 	 {{M}entor{N}et: Learning Data-Driven Curriculum for Very Deep Neural Networks on Corrupted Labels},
  author =       {Jiang, Lu and Zhou, Zhengyuan and Leung, Thomas and Li, Li-Jia and Fei-Fei, Li},
  booktitle = 	 {Proceedings of the 35th International Conference on Machine Learning},
  pages = 	 {2304--2313},
  year = 	 {2018},
  volume = 	 {80},
  series = 	 {Proceedings of Machine Learning Research},
  month = 	 {10--15 Jul}
}

@inproceedings{Han2018co,
 author = {Han, Bo and Yao, Quanming and Yu, Xingrui and Niu, Gang and Xu, Miao and Hu, Weihua and Tsang, Ivor and Sugiyama, Masashi},
 booktitle = {Advances in Neural Information Processing Systems},
 pages = {},
 title = {Co-teaching: Robust training of deep neural networks with extremely noisy labels},
 volume = {31},
 year = {2018}
}

@inproceedings{Eran2017Decoupling,
 author = {Malach, Eran and Shalev-Shwartz, Shai},
 booktitle = {Advances in Neural Information Processing Systems},
 pages = {},
 title = {Decoupling "when to update" from "how to update"},
 volume = {30},
 year = {2017}
}

@InProceedings{Yu2019How,
  title = 	 {How does Disagreement Help Generalization against Label Corruption?},
  author =       {Yu, Xingrui and Han, Bo and Yao, Jiangchao and Niu, Gang and Tsang, Ivor and Sugiyama, Masashi},
  booktitle = 	 {Proceedings of the 36th International Conference on Machine Learning},
  pages = 	 {7164--7173},
  year = 	 {2019},
  volume = 	 {97},
  series = 	 {Proceedings of Machine Learning Research},
  month = 	 {09--15 Jun}
}

@inproceedings{Xu2019LDMI,
 author = {Xu, Yilun and Cao, Peng and Kong, Yuqing and Wang, Yizhou},
 booktitle = {Advances in Neural Information Processing Systems},
 pages = {},
 title = {L\textsubscript{DMI}: A Novel Information-theoretic Loss Function for Training Deep Nets Robust to Label Noise},
 volume = {32},
 year = {2019}
}

@article{Anurag2019SeCoST,
        title        = {SeCoST: Sequential Co-Supervision for Weakly Labeled Audio Event Detection},
      author       = {Anurag Kumar and Vamsi Krishna Ithapu},
    journal      = {arXiv preprint arXiv:1910.11789},
      year={2019},
     eprinttype    = {arXiv},
      eprint={1910.11789}
}

@INPROCEEDINGS{Lee2018CleanNet,
  author={Lee, Kuang-Huei and He, Xiaodong and Zhang, Lei and Yang, Linjun},
  booktitle={2018 IEEE/CVF Conference on Computer Vision and Pattern Recognition}, 
  title={CleanNet: Transfer Learning for Scalable Image Classifier Training with Label Noise}, 
  year={2018},
  volume={},
  number={},
  pages={5447-5456}
}

@inproceedings{Vahdat2017Toward,
 author = {Vahdat, Arash},
 booktitle = {Advances in Neural Information Processing Systems},
 pages = {},
 publisher = {Curran Associates, Inc.},
 title = {Toward Robustness against Label Noise in Training Deep Discriminative Neural Networks},
 volume = {30},
 year = {2017}
}

@INPROCEEDINGS{Li2017Learning,
  author={Li, Yuncheng and Yang, Jianchao and Song, Yale and Cao, Liangliang and Luo, Jiebo and Li, Li-Jia},
  booktitle={2017 IEEE International Conference on Computer Vision (ICCV)}, 
  title={Learning from Noisy Labels with Distillation}, 
  year={2017},
  volume={},
  number={},
  pages={1928-1936}
}

@INPROCEEDINGS{Veit2017Learninig,
  author={Veit, Andreas and Alldrin, Neil and Chechik, Gal and Krasin, Ivan and Gupta, Abhinav and Belongie, Serge},
  booktitle={2017 IEEE Conference on Computer Vision and Pattern Recognition (CVPR)}, 
  title={Learning from Noisy Large-Scale Datasets with Minimal Supervision}, 
  year={2017},
  volume={},
  number={},
  pages={6575-6583}
}

@InProceedings{Giorgio2017Making,
author = {Patrini, Giorgio and Rozza, Alessandro and Krishna Menon, Aditya and Nock, Richard and Qu, Lizhen},
title = {Making Deep Neural Networks Robust to Label Noise: A Loss Correction Approach},
booktitle = {Proceedings of the IEEE Conference on Computer Vision and Pattern Recognition (CVPR)},
month = {July},
year = {2017}
}

@inproceedings{Han2018Masking,
 author = {Han, Bo and Yao, Jiangchao and Niu, Gang and Zhou, Mingyuan and Tsang, Ivor and Zhang, Ya and Sugiyama, Masashi},
 booktitle = {Advances in Neural Information Processing Systems},
 pages = {},
 title = {Masking: A New Perspective of Noisy Supervision},
 volume = {31},
 year = {2018}
}

@INPROCEEDINGS{Szegedy2016Rethinking,
  author={Szegedy, Christian and Vanhoucke, Vincent and Ioffe, Sergey and Shlens, Jon and Wojna, Zbigniew},
  booktitle={2016 IEEE Conference on Computer Vision and Pattern Recognition (CVPR)}, 
  title={Rethinking the Inception Architecture for Computer Vision}, 
  year={2016},
  volume={},
  number={},
  pages={2818-2826}
}

@inproceedings{goldberger2017training,
title={Training deep neural-networks using a noise adaptation layer},
author={Jacob Goldberger and Ehud Ben-Reuven},
booktitle={International Conference on Learning Representations},
year={2017}
}

\end{document}